\documentclass[preprint,12pt,authoryear]{elsarticle}

\usepackage{amssymb}
\usepackage{graphicx, array, lscape, rotating, amsmath, comment, adjustbox, booktabs, amsfonts, algorithm, multirow, makecell}
\usepackage{algorithmic}
\usepackage[referable]{threeparttablex}
\usepackage[T1]{fontenc}
\newcommand{\NA}{---}

\journal{Neural Networks}

\begin{document}

\begin{frontmatter}



\title{Reinforcement learning for automatic quadrilateral mesh generation: a soft actor-critic approach}


\author[inst1]{Jie Pan}

\affiliation[inst1]{organization={Concordia Institute for Information Systems Engineering, Concordia University},
            city={Montreal},
            postcode={H3G 1M8}, 
            state={Quebec},
            country={Canada}}

\author[inst2]{Jingwei Huang}

\affiliation[inst2]{organization={Department of Engineering Management \& Systems Engineering, Old Dominion University},
            city={Norfolk},
            postcode={23529}, 
            state={Virginia},
            country={United States}}
            
\author[inst3]{Gengdong Cheng}

\affiliation[inst3]{organization={Department of Engineering Mechanics, Dalian University of Technology},
            city={Dalian},
            postcode={116023}, 
            state={Liaoning},
            country={China}}

\author[inst1]{Yong Zeng \corref{cor1}}
\ead{yong.zeng@concordia.ca}
\cortext[cor1]{Corresponding author}

\begin{abstract}
This paper proposes, implements, and evaluates a reinforcement learning (RL)-based computational framework for automatic mesh generation. Mesh generation plays a fundamental role in numerical simulations in the area of computer aided design and engineering (CAD/E). It is identified as one of the critical issues in the NASA CFD Vision 2030 Study. Existing mesh generation methods suffer from high computational complexity, low mesh quality in complex geometries, and speed limitations. These methods and tools, including commercial software packages, are typically semiautomatic and they need inputs or help from human experts. By formulating the mesh generation as a Markov decision process (MDP) problem, we are able to use a state-of-the-art reinforcement learning (RL) algorithm called “soft actor-critic” to automatically learn from trials the policy of actions for mesh generation. The implementation of this RL algorithm for mesh generation allows us to build a fully automatic mesh generation system without human intervention and any extra clean-up operations, which fills the gap in the existing mesh generation tools. In the experiments to compare with two representative commercial software packages, our system demonstrates promising performance with respect to scalability, generalizability, and effectiveness.
\end{abstract}



\end{frontmatter}


\section{Introduction}
\label{sec:introduction}


Reinforcement learning (RL) has been researched and applied in many fields, such as games \citep{silver_mastering_2016}, health care \citep{gottesman2019guidelines}, natural language processing \citep{wang2018deep}, and combinatorial optimization, particularly for NP-hard problems \citep{mazyavkina2021reinforcement}. However, it has rarely been applied to the area of computational geometry, especially in the field of mesh generation \citep{pan_huang_wang_cheng_zeng_2021}. Mesh generation is a fundamental step in conducting numerical simulations in the area of Computer-Aided Design and Engineering (CAD/E) such as finite element analysis (FEA), computational fluid dynamics (CFD), and graphic model rendering \citep{gordon_construction_1973, Yao2019}. Mesh generation techniques have been identified as one of the six basic capacities to build in NASA's Vision 2030 CFD Study \citep{slotnick2014cfd}. 

\begin{figure}
    \centering
    \includegraphics[width=0.8\textwidth]{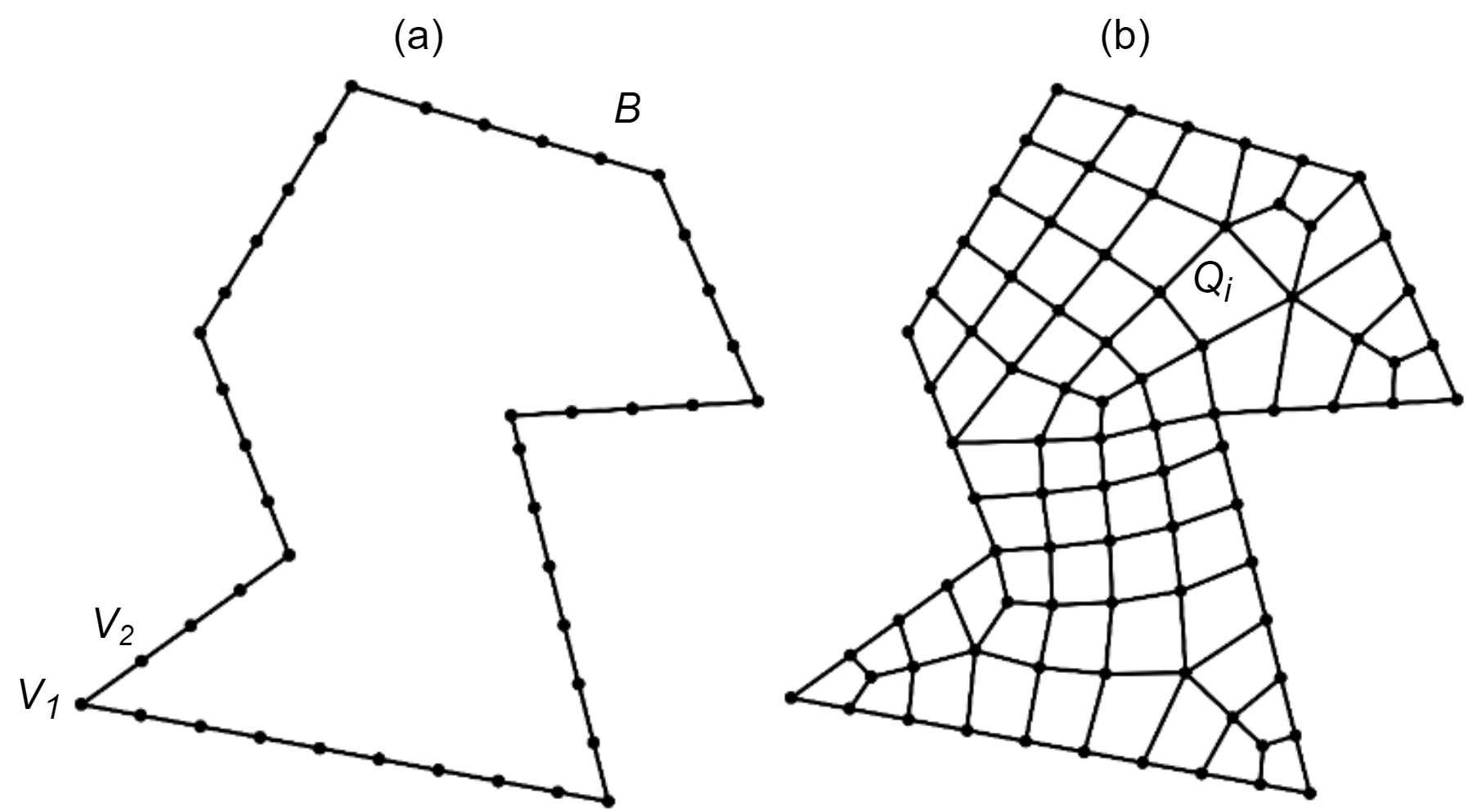}
    \caption{Meshing problem. (a) The initial geometric boundary, $B$, is composed of piecewise linear segments, represented as a sequence of vertices $[V_1, V_2, ..., V_N]$; (b) The final mesh,  $\Omega$, is composed of a set of quadrilateral elements $[Q_1, Q_2, ..., Q_M]$.}
    \label{fig:meshing problem}
\end{figure}

Mesh generation discretizes a complex geometric domain (see Fig. \ref{fig:meshing problem} (a)) into a finite set of (geometrically simple and bounded) elements, such as 2D triangles or quadrilaterals (see Fig. \ref{fig:meshing problem} (b) in 2D geometries) and tetrahedra or hexahedra (in 3D geometries). Since the reliable automation and high quality of mesh representation matter significantly to the numerical simulation results, mesh generation has continued to be a significant bottleneck in those fields due to algorithm complexities, inadequate error estimation capabilities, and complex geometries \citep{slotnick2014cfd}. 

Strongly favored by engineering analysis, quadrilateral mesh generation has been a critical research problem for decades. However, existing quadrilateral mesh generation methods require heuristic knowledge in algorithm development and severely depend on preprocessing or postprocessing operations to maintain high mesh quality. Preprocessing includes decomposing complex domains into multiple regular components \citep{liu2017distributed} and generating favorable vertex locations \citep{remacle_frontal_2013}. Postprocessing (i.e., clean-up operations) is used to handle where the mesh consists of elements other than quadrilaterals (e.g., triangular elements), flat or inverted elements, and irregularly connected elements. The operations include reducing the singularity \citep{verma_robust_2017}, performing iterative topological changes (e.g., splitting, swapping, and collapsing elements) \citep{docampo-sanchez_towards_2019}, and mesh adaptation \citep{verma2018alphamst}. 

Improving mesh quality with extra operations increases computational complexity and slows the meshing speed. Many machine learning-based methods are proposed to support mesh optimization and reconstruction \citep{zhang_meshingnet_2020, yang2021reinforcement, gupta2020neural, chen2018neural, wang2018pixel2mesh, defferrard2016convolutional}. \citet{nechaeva_composite_2006} proposed an adaptive mesh generation algorithm based on self-organizing maps (SOMs), which adapts a given uniform mesh onto a target physical domain through mapping. Pointer networks can generate triangular meshes, but the result is not robust and contains intersecting edges \citep{vinyals_pointer_2015}. \citet{papagiannopoulos_how_2021} proposed a triangular mesh generation method with three neural networks. The training data came from the constrained Delaunay triangulation algorithm \citep{chew_constrained_1989}. The trained model could predict the number of candidate inner vertices (to form a triangular element), the coordinates of those vertices, and their connective relations with existing segments on the boundary. Adapting to arbitrary and complex geometries is a shortcoming because of the fixed input scale and constrained diversity of training samples.

To achieve generalizability for arbitrarily shaped geometries, \citet{zeng_knowledge-based_1993} used a recursive algorithm to generate one element at a time by three primitive rules to avoid taking the whole contour as the input. \citet{yao_ann-based_2005} improved the recursive method by using an artificial neural network (ANN) to automate the rule learning. However, the quality and diversity of training data cannot be achieved and limit the model performance.  

\begin{figure*}
    \centering
    \includegraphics[width=\textwidth]{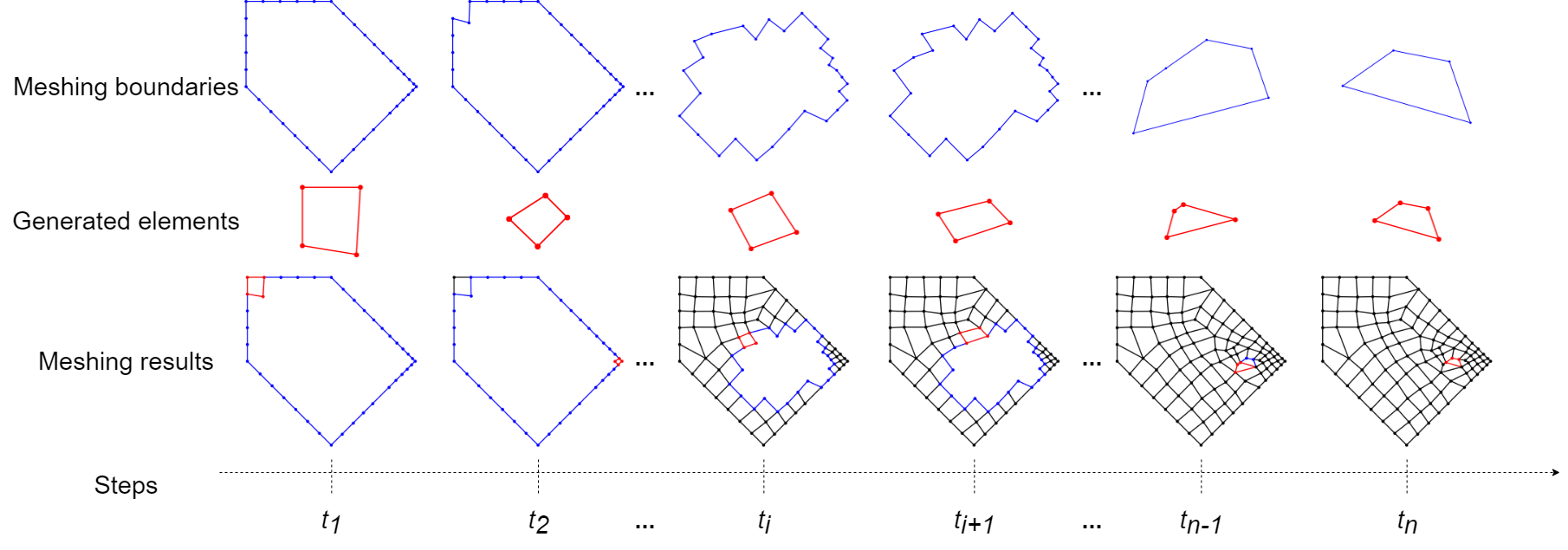}
    \caption{A sequence of actions taken by the mesh generator to complete a mesh. At each time step $t_i$, an element (in red) is extracted from the current boundary (in blue). The boundary is then updated by cutting off the element and serves as the meshing boundary in the next time step $t_{i+1}$. This process continues until the updated boundary becomes an element.}
    \label{fig:rl mesh}
\end{figure*}

To resolve this issue, \citet{pan_huang_wang_cheng_zeng_2021} formulated the mesh generation algorithm by \citet{zeng_knowledge-based_1993} into a Markov decision problem (MDP). The formalized problem framework was addressed by reinforcement learning \citep{sutton_reinforcement_1998,sutton_reinforcement_2018}, as illustrated in Fig. \ref{fig:rl mesh}. At each time step, a quadrilateral element is generated from the domain boundary. The boundary is inwardly updated by cutting off the generated element. In each iteration, the meshing boundary evolves into a new boundary to generate new elements, which is naturally a sequential decision-making process. In the previous work \citep{pan_huang_wang_cheng_zeng_2021}, the present authors only used the RL method to sample training data for a feedforward neural network model, which did not fully exploit the potential of the RL models and resulted in an extra sampling selection phase. This heuristic selection can easily cause an imbalanced dataset and compromise model performance.

\begin{figure}
    \centering
    \includegraphics[width=0.8\textwidth]{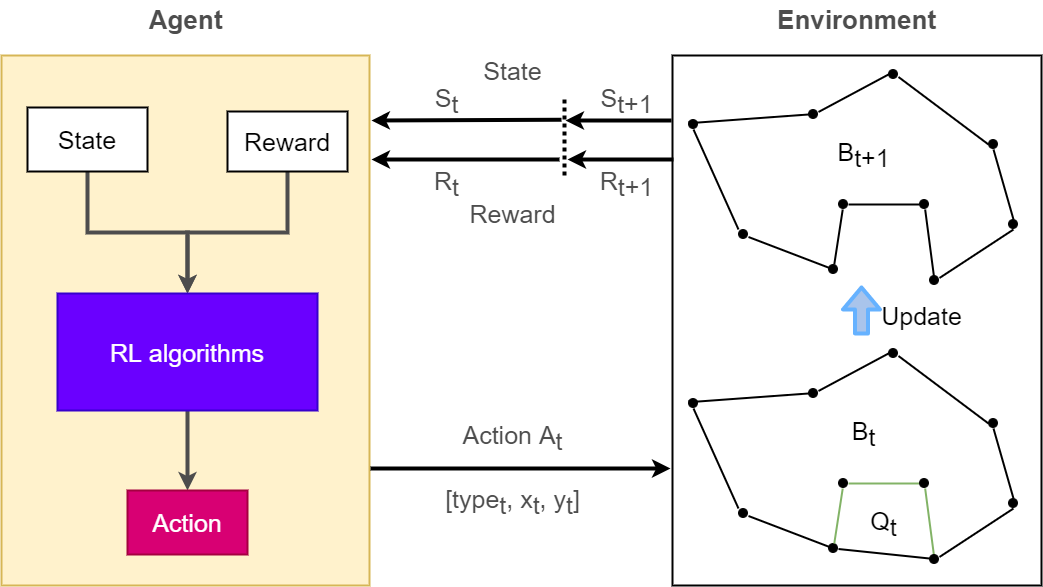}
    \caption{The RL-based computational framework for automatic mesh generation. The agent acts as the mesh generator by implementing various RL techniques. It generates an element after a state is perceived and improves the action by observing the rewards. The environment models the meshing boundary and updates the boundary by cutting off the element generated via the action.}
    \label{fig:Mesh RL architecture}
\end{figure}

The present article resolves the abovementioned problem and provides an RL-based computational framework, FreeMesh-RL, for quadrilateral mesh generation that automatically collects balanced samples through trial-and-error learning, as illustrated in Fig. \ref{fig:Mesh RL architecture}. It aims to automatically provide high-quality meshes for various complex geometries without human intervention. A few challenges must be addressed: 1) the RL method needs to be robust and less hyperparameter-sensitive; 2) the trade-off between the current element quality and the remaining boundary quality shall be made to maintain overall high mesh quality; and 3) the meshing process should be finished in finite steps. Compared with other 
RL algorithms \citep{lillicrap2015continuous, mnih_asynchronous_2016, schulman2017proximal, fujimoto2018addressing}, soft actor-critic (SAC) \citep{haarnoja2018soft, haarnoja2018soft_applications} is 
selected
to address the mesh generation problem, because of its 
capability allowing reuse of previous experience, 
good balance between exploration and exploitation, 
stable learning efficiency, and being less hyperparameter-sensitive.

The rest of the present paper is organized as follows. Section \ref{sec: method} introduces the detailed formulation of mesh generation as an RL problem, including action, state, and reward, and proposes an RL-based meshing architecture with SAC. Section \ref{sec: experiment} explores the implementation details of each key concepts in applying SAC to mesh generation, and evaluates the performance of the proposed method in comparison with two state-of-the-art meshing approaches that share the similar algorithmic strategy yet with the rules and knowledge developed based on expert experience. Section \ref{sec: discussion} discusses the main findings based on the experiment results and explains how the method can be applied to a broader area both in RL and mesh generation. Section \ref{sec: conclusion} concludes this article.

\section{RL based mesh generation}
\label{sec: method}
This section proposes an RL-based computational framework for automatic mesh generation. The action formulation and state representation are detailed first. A reward function is then designed to meet mesh quality requirements. Finally, the architecture of the proposed framework is explained.

\subsection{Action formulation}
In the mesh generation process (see Fig. \ref{fig:rl mesh}), two important decisions should be made: 1) how to choose a vertex (called the reference vertex in this paper) from the boundary; and 2) how to decide on the other three vertices to form a quadrilateral element.

\begin{figure}
    \centering
    \includegraphics[width=0.5\textwidth]{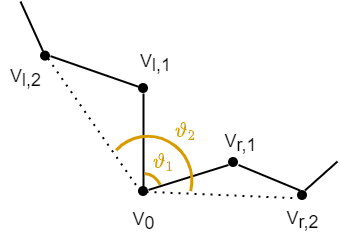}
    \caption{The angle calculation in reference vertex selection. For example, two surrounding vertices of $V_0$, $n_{rv} = 2$, are used to calculate the angles $\vartheta_1$ and $\vartheta_2$, where $\vartheta_1 = \angle V_{l,1}V_0 V_{r,1}$, $\vartheta_2 = \angle V_{l,2}V_0 V_{r,2}$. This calculation iterates along the boundary. The vertex with the least averaged angle by $\vartheta_1$ and $\vartheta_2$ is selected as the reference vertex.}
    \label{fig: reference_vertex}
\end{figure}

The first decision is to select a local reference region that has the least boundary angle from the boundary. It reduces the formation of narrow regions in the remaining boundary, and thus increases high-quality elements generated by subsequent actions. The reference vertex $V_i^*$ is selected using the following equation:
\begin{equation}
\label{eqn: reference vertex}
V_i^* = \mathop{\arg \min} \limits_{V_i} \frac{1}{n_{rv}} \sum_{j=1}^{n_{rv}}{\angle V_{l,j}V_i V_{r,j}}, i \in N_B,
\end{equation}
where $N_B$ is the number of vertices contained in the present boundary $B$; $V_{l,j}$ and $V_{r,j}$ denote the $j$-th vertices at the left and right side (considering clockwise orientation) of the reference vertex $V_i$ along the boundary, respectively; $n_{rv}$ represents how many surrounding vertices should be included; the $V_i$ is the $i$-th vertex on the boundary. An example of this selection is shown in Fig. \ref{fig: reference_vertex}. The detailed selection is discussed in the experiment section.

\begin{figure}
    \centering
    \includegraphics[width=0.8\textwidth]{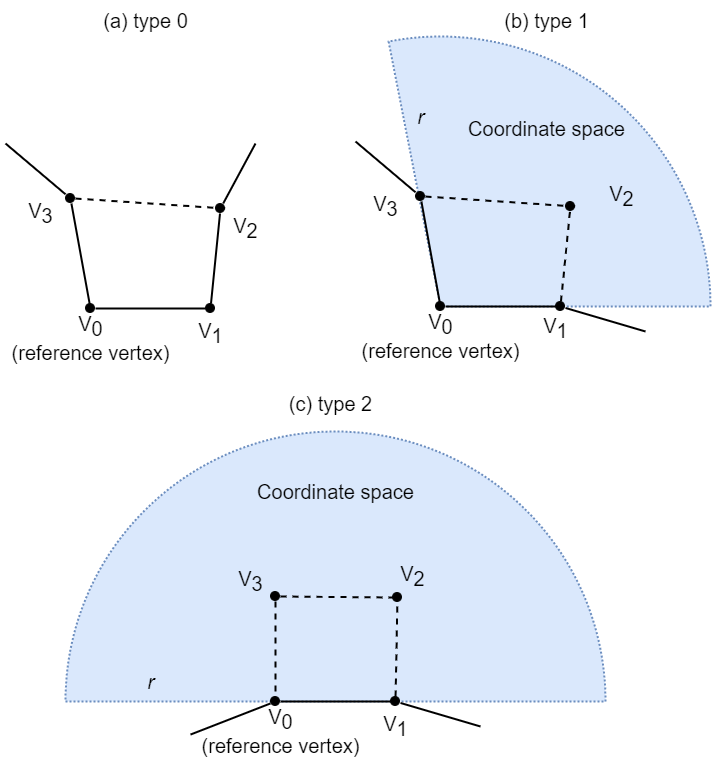}
    \caption{Action space for each rule type. Subfigures (a)-(c) correspond to three types of actions, respectively. The blue area is the space to select the candidate vertices with a radius $r$ and the origin vertex $V_0$ (reference vertex). The candidate vertices are $V_2$ (in type 1), and $V_2$ and $V_3$ (in type 2), respectively.}
    \label{fig:action space}
\end{figure}

The second decision is to choose an correct action to form a quadrilateral element from three basic action types: adding zero, one, or two more vertices \citep{zeng_knowledge-based_1993}. As shown in Fig. \ref{fig:action space}, the action is represented by $[type, V_1, V_2]$, where $type \in \{0, 1, 2\}$, corresponding to the three possible actions; $V_1$ and $V_2$ are the coordinates of the newly added vertices. The coordinate space for the vertices is constrained to a fan-shaped area (in light blue) with radius $r$, which is calculated as follows:
\begin{equation}
\label{eqn: radius}
\begin{split}
    r &=\alpha * L, \\
    L &= \frac{1}{2n}\sum_{j=0}^n{|V_{l,j}V_{l,j+1}}| + |V_{r,j}V_{r,j+1}|, 0 < n < N/2 \\
    \end{split}
\end{equation}
where $\alpha$ is a factor to amplify the base length $L$; $V_{l,j}$ and $V_{r,j}$ denote the $j$-th vertex at the left and right side of the reference vertex along the boundary; $V_{l,0} = V_i = V_{r,0}$ is the reference vertex; and $|V_aV_b|$ is the Euclidean distance between two vertices. Usually, the action $type=2$ is needed only on special occasions (e.g., circular domains) at limited times. It is implemented in the environment if these situations occur. The action $[type, V_1]$ is eventually used in this article.

\subsection{State representation}
A state is an observation of the environment by the agent. The environment here is the meshing boundary (see Fig. \ref{fig:rl mesh}). The full information for the state of the environment at time $t$ consists of all the vertices along the boundary. However, not all the vertices are relevant and necessary to deciding what action to take. Therefore, a partial observation of the boundary environment is designed, which consists of a reference vertex determining where the agent should start generating an element, and its surrounding vertices providing a local environment.

\begin{figure}
    \centering
    \includegraphics[width=0.4\textwidth]{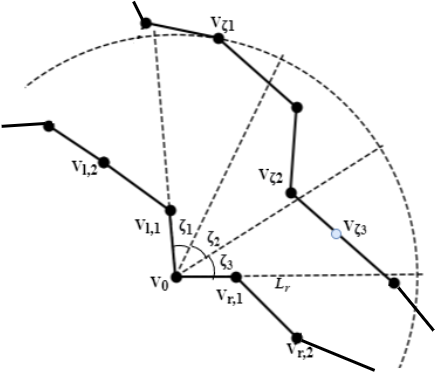}
    \caption{Partial observation of the meshing boundary. For example, the partial boundary, where $\beta=4, n=2, g=3$, is represented as the state. First, two vertices on the left and right sides of the reference vertex $V_0$ are selected. Second, the angle $\angle V_{l,1}V_0V_{r,1}$ is evenly split into three angles $\zeta_1, \zeta_2, $ and $\zeta_3$; three fan-shaped areas are hence formed with these angles and a radius $L_r=4 * L$. Then, the closest vertex in each area is selected. $V_{\zeta_3}$ is an intersection vertex between the bisector and the boundary segment.}
    \label{fig:paritial boundary}
\end{figure}

A state at time $t$ is denoted as $s_t$, as shown in Fig. \ref{fig:paritial boundary}, and is composed of five components: (1) a reference vertex, $V_0$, which is used as the relative origin to generate the new element with an action $a_t$, and is calculated by Equation \ref{eqn: reference vertex}; (2) $n$ neighboring vertices on the right side of the reference vertex; (3) $n$ neighboring vertices on the left side; (4) $g$ neighboring points $V_{\zeta_1}, ..., V_{\zeta_g}$, the closest vertices in the fan-shaped area $\zeta_1, ..., \zeta_g$ within a radius, which is calculated as:
\begin{equation}
\label{eqn: fan radius}
   L_r=\beta * L,
\end{equation}
where $\beta$ is a factor to amplify the base length and $L$ is calculated by Equation \ref{eqn: radius}. When $g=3$, the fan-shaped area is evenly divided into three parts, $\zeta_1 =\zeta_2 = \zeta_3$, as shown in Fig. \ref{fig:paritial boundary}. If there are no vertices in a sliced fan-shaped area (e.g., $\zeta_3$), the furthest bisector vertex in the slice or the intersection vertex between the bisector of the slice and the boundary edge is selected, such as $V_{\zeta_3}$ in Fig. \ref{fig:paritial boundary}; and (5) $\rho_t$, the area ratio between the updated domain and the original domain, which is used to indicate the meshing progress.

The current state $S_t$, representing the partial boundary around the reference point and meshing progress, is denoted as follows:
\begin{equation}
\label{eqn: state function}
    S_t=\{V_{l,n}, ..., V_{l,1}, V_0, V_{r,1}, ..., V_{r,n}, V_{\zeta_1}, ..., V_{\zeta_g}, \rho_t\}.
\end{equation}
All the vertices are represented by a polar coordinate system with $V_0$ as the origin and $\overrightarrow{V_0V_{r,1}}$ as the reference direction. There are a few geometrical operations (i.e., rotation, scaling, and transit) to keep only the relative information of the contained vertices \citep{yao_ann-based_2005}.

\subsection{Reward function}
The criteria for high-quality meshing are as follows: 1) each element is a quadrilateral; (2) each element should be as close to a square as possible, and minimally, the inner corners of each element should be between 45° and 135°; (3) the aspect ratio (the ratio of opposite edges) and taper ratio (the ratio of neighboring edges) of each quadrilateral should be within a predefined range; and (4) the transition between a dense mesh and a coarse mesh should be smooth \citep{zeng_knowledge-based_1993, zeng_understanding_2009}.

The reward function shall guarantee the quality requirements and step-wisely measure the performance of each action. The action can cause three situations: 1) if it forms an invalid element or intersects with the boundary, the reward is set to -0.1; 2) if it generates the last element, the reward is set to 10; and 3) if it constructs a valid element, the reward is a joint measurement of element quality, the quality of the remaining boundary, and the density. Consequently, the reward function is represented as follows:

\begin{equation} 
\label{reward function}
 r_t(s_t, a_t) = \begin{cases}
 -0.1, & \text{invalid element}; \\
 10, & \text{the element is the last element}; \\
 m_t, & \text{otherwise}. \\
 \end{cases}
\end{equation}

The measurement $m_t$ is calculated by the following equation:
\begin{equation}
    m_t=\eta^e_t + \eta^b_t + \mu_t.
\end{equation}

\begin{figure}
    \centering
    \includegraphics[width=0.8\textwidth]{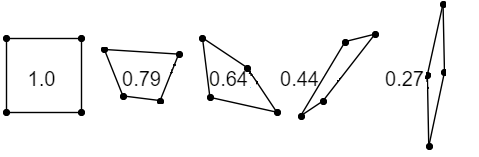}
    \caption{Element quality varies with different shapes. The quality value ranges from 0 to 1. The element with the best quality 1 is a square.}
    \label{fig: elements quality}
\end{figure}

The element quality $\eta^e_t$ is measured by its edges and internal angles, and is calculated as follows:

\begin{equation} 
\label{element quality}
\begin{split}
 \eta^e_t & = \sqrt{q^{edge} q^{angle}}, \\
q^{edge} & = \frac{\sqrt{2} min_{j\in\{0,1,2,3\}}\{l_j\}}{D_{max}}, \\
q^{angle} & = \frac{min_{j\in\{0,1,2,3\}}\{angle_j\}}{max_{j\in\{0,1,2,3\}}\{angle_j\}},
\end{split}
\end{equation}
where $q^{edge}$ refers to the quality of edges of this element; $l_j$ is the length of the $j$th edge of the element; $D_{max}$ is the length of the longest diagonal of the $t$th element; $q^{angle}$ refers to the quality of the angles of the element; and $angle_j$ is the degree of the $j$th inner angle of the element. The quality $\eta_t^e$ will range from 0 to 1, which is the greater the better. Examples of various element qualities are shown in Fig. \ref{fig: elements quality}.

\begin{figure}
    \centering
    \includegraphics[width=0.8\textwidth]{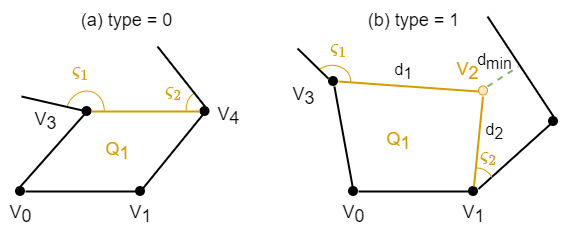}
    \caption{The quality of the remaining boundary for two action types. $Q_1$ is the newly generated element. Once it is removed, it forms two angles $\varsigma_1$ and $\varsigma_2$ with existing boundary segments. (a) When action type = 0, the boundary quality only hinges on these two angles, assuming $q^{dist} = 1$. (b) When action type = 1, the boundary quality is jointly measured by the two angles, and the closest Euclidean distance $d_{min}$ of the newly added vertex $V_2$ to the existing segments. $d_1$ and $d_2$ are the Euclidean lengths of segments $V_2V_3$ and $V_1V_2$.}
    \label{fig: boundary quality}
\end{figure}

The quality of the remaining boundary $\eta^b_t$ is measured by both the quality of the angles formed between the newly generated element and the boundary, and the shortest distance of the generated vertex to its surrounding edges, and is denoted as follows:
\begin{equation}
\begin{split}
    \eta^b_t &= \sqrt{\frac{min_{k\in\{1,2\}}\{min(\varsigma_k, M_{angle})\}}{M_{angle}} q^{dist}} - 1, \\
    q^{dist} &= \begin{cases}
   \frac{d_{min}}{(d_1 + d_2)/2}, & \text{if } d_{min} < (d_1 + d_2)/2; \\
   1, & \text{otherwise. } \\
 \end{cases}
\end{split}
\end{equation}
where $\varsigma_k$ refers to the degrees of the $k$th generated angle and $d_{min}$ is the distance of $V_2$ to its closest edge. The details are shown in Fig. \ref{fig: boundary quality}. The quality $\eta_t^b$ ranges from -1 to 0, with a larger value representing better quality. It serves as a penalty term to decrease the reward if the quality of the remaining boundary worsens. We set $M_{angle}$ = $60^{\circ}$. When the formed new angles are less than $M_{angle}$, the quality will decrease. It penalizes the generation of sharp angles that are harmful to the overall mesh quality and may even fail the meshing process.

These two qualities together represent the trade-off between the generated element and the remaining boundary, ensuring the overall mesh quality. The last term, density $\mu_t$, secures the completion of the meshing process within finite steps by controlling mesh density, which is calculated as follows:

\begin{equation} 
\label{density factor}
 \mu_t = \begin{cases}
   -1, & \text{if }\mathcal{A}_t < \mathcal{A}_{min}; \\
 \frac{\mathcal{A}_t - \mathcal{A}_{min}}{\mathcal{A}_{max} - \mathcal{A}_{min}}, & \text{if }\mathcal{A}_{min} \leq \mathcal{A}_t < \mathcal{A}_{max} ; \\
 0, & \text{otherwise.} \\
 \end{cases}
\end{equation}
where $\mathcal{A}_t$ is the area of the element generated at time $t$; $\mathcal{A}_{min}$ is the estimated minimum area that an element should meet, and is calculated by $\mathcal{A}_{min} = \upsilon \cdot e_{min}^2$; $\mathcal{A}_{max}$ is the estimated maximum area of the element, and is calculated by $\mathcal{A}_{max} = \upsilon \left(\frac{e_{max}-e_{min}}{\kappa} + e_{min}\right)^2$; $e_{max}$ and $e_{min}$ are the lengths of the longest and shortest edges in the boundary, respectively; ${\kappa}$ adjusts the estimated maximum area and is independent of the unit of edges (${\kappa} = 4$ in our experiments); and $\upsilon$ is a weight and values in $(0, 10]$, 
for which a smaller value means a greater density, and vice versa. We set $\upsilon = 1$ in our experiments for the medium density.

\subsection{Meshing scheme via SAC}
The formulated RL-based meshing architecture, FreeMesh-RL, is shown in Fig. \ref{fig:Mesh RL architecture}. We formulate meshing process as an MDP process, consisting of a set of boundary environment states $\mathcal{S}$, a set of possible actions $\mathcal{A}(s)$, a set of rewards $\mathcal{R}$, and a state transition probability $P(S_{t+1}, R_{t+1}|S_t, A_t)$. The agent, at each time step t, observes a state $S_t$ from the environment, and conducts an action $A_t$ applied to the environment. The environment responds to the action and transitions into a new state $S_{t+1}$. It then reveals the new state and provides a reward $R_t$ to the agent. This process iterates until a given condition is satisfied (i.e., the RL problem is solved). The extraction process will produce a sequence $[S_0, A_0, S_1, R_1, A_1, ...]$. The goal of the meshing agent is to complete the meshing process for any given geometric object while maintaining high mesh quality. 

The RL algorithm deployed in this computational framework is the SAC method. SAC is one of the state-of-the-art RL algorithms for continuous action control problems \citep{haarnoja2018soft, haarnoja2018soft_applications}. To overcome the sample complexity and hyperparameter-sensitivity, it adds an entropy term in addition to the reward in the objective function, and maximizes the reward return while maximizing the randomness of the policy. 

Following \citep{haarnoja2018soft}, 
the objective function of the policy is correspondingly denoted as 
\begin{equation}
    J(\pi)= \sum_{t=0}^{T}\mathbb{E}_{(s_t,a_t)\sim\rho_{\pi_\theta}}[r(s_t, a_t) + \alpha \mathcal{H}(\pi_\theta(.\vert{s_t}))],
\end{equation}
where $\mathcal{H}(.)$ is the entropy measure \citep{ziebart2010modeling}; $\rho_{\pi_\theta}$ is the state-action marginal distribution of policy $\pi$ parameterized by $\theta$; and $\alpha$ indicates the significance of the entropy term, known as the temperature parameter. Entropy maximization allows the learned policy to act as randomly as possible while guaranteeing task completion, which gains a trade-off between exploration and exploitation and thus accelerates learning. This randomness is especially important for a partially observable environment.

SAC combines Q-learning with stochastic policy gradient learning. Q-learning is based on the Bellman equation
\begin{equation}
    Q(s_t, a_t) = r(s_t, a_t) + \gamma \mathbb{E}_{s_{t+1} \sim P} (\max_{a'} Q(s_{t+1}, a')).
\end{equation}
As we already know, in SAC, policy entropy maximization is introduced in the RL objective of maximizing the expected return. Correspondingly, we have the following form of the soft Bellman equation.
\begin{equation}
    Q(s_t, a_t)=r(s_t, a_t)+\gamma \mathbb{E}_{s_{t+1}\sim \rho_\pi(s)} [V(s_{t+1})], 
\end{equation}
and 
\begin{equation}
    V(s_t)=\mathbb{E}_{a_t\sim \pi}[Q(s_t, a_t)-\alpha \mathrm{log}\pi(a_t\vert{s_t})], 
\end{equation}
where $\rho_\pi(s)$ is the state marginals of the trajectory distribution induced by a policy $\pi(a_t\vert{s_t})$. 

SAC uses this soft Bellman equation to estimate the target soft Q-values, and soft Q-learning in SAC aims to minimize the difference between the Q-value estimated by a Q-function approximator with parameters $\theta$ and the target soft Q-value as follows.
\begin{equation}
J_Q(\theta) = \mathbb{E}_{s_t,a_t\sim \mathcal{D}}\left[\frac{1}{2} (Q_\theta(s_t, a_t) - \hat{Q}(s_t, a_t))^2\right].
\end{equation}
In training, all samples are taken from the replay buffer $\mathcal{D}$, which is the collection of previous experience, i.e. $(s,a,r,s')$ tuples. SAC uses neural networks to approximate the soft Q-function and the policy.

To train Q-function neural network with parameters $\theta$ by using gradient descendent, we can estimate the gradient as follows, 
\begin{multline}
\hat{\nabla}_\theta J_Q(\theta) = \nabla_\theta Q_\theta(s_t, a_t)(Q_\theta (s_t, a_t) - \\
(r(s_t, a_t) + 
\gamma (Q_{\bar{\theta}} (s_{t+1}, a_{t+1})-\alpha \mathrm{log} (\pi_\phi (a_{t+1}\vert{s_{t+1}})))),
\end{multline}
where $\pi_\phi$ is the current policy parameterized by a neural network with parameters $\phi$.
To simplify gradient descent, the target Q-value is calculated with a different neural network with parameters $\bar{\theta}$, which is simply the corresponding soft Q-function network with delayed parameter updates. Practically,
$\bar{\theta}_i$ is obtained as an exponential moving average of $\theta_i$. 

As in the TD3 (Twin Delayed DDPG) model \citep{fujimoto2018addressing}, SAC has a neural network with parameters $\phi$ to approximate policy, two neural networks, each with parameters $\theta_i$ ($i = 1, 2$), working together to approximate the soft Q-function, and two neural networks, each with parameters $\bar{\theta}_i$ ($i = 1, 2$), to estimate target Q-values. The latter two neural networks are simply the former two networks with the delayed updates of parameters $\theta_i$. 
In order to overcome the overestimation bias \citep{fujimoto2018addressing}, the minimum of the two values estimated by the twin networks is used as the estimated Q-value, i.e., 
$Q_\theta(s_t, a_t) = \min \{Q_{\theta_1}(s_t, a_t), Q_{\theta_2}(s_t, a_t)\}$.

In the soft policy improvement stage, 
following \citep{haarnoja2018soft},
the policy parameter can be updated by minimizing the expected Kullback-Leibler (KL)-divergence.
The learning objective can be expressed as follows,
\begin{equation}
    J(\phi)=\mathbb{E}_{s_t\sim \mathcal{D}}[\mathbb{E}_{a_t \sim \pi_\phi}[\alpha \mathrm{log} (\pi_\phi (a_t\vert{s_t})) - Q_\theta (s_t, a_t)]].
\end{equation}
The policy is reparamerterized with a neural network transformation 
    $a_t = f_\phi(\epsilon_t; s_t)$,
where $\epsilon_t$ is an input noise vector and can be sampled from a fixed distribution.

Finally, the temperature $\alpha$ is updated by minimizing the objective 
\begin{equation}
    J(\alpha)=\mathbb{E}_{a_t \sim \pi}[-\alpha \mathrm{log} \pi_\phi (a_t\vert{s_t})-\alpha \bar{\mathcal{H}}].
\end{equation}
The details of the algorithm are shown in Algorithm \ref{alg: sac}. The parameters used are detailed in the experiment section.

\begin{algorithm}[h]
\caption{SAC for mesh generation.}
 \label{alg: sac}
\begin{algorithmic}[1]
\STATE Initialize parameters for Q networks and policy network, $\theta_1, \theta_2 $, $\phi$; initialize replay buffer $\mathcal{D}$; initialize environment \;
\STATE Set target Q networks $\bar{\theta}_1 \leftarrow \theta_1, \bar{\theta}_2 \leftarrow \theta_2$ \;
\FOR{time step $t$ in range (1, $N_T$), }
\STATE Select action $a_t \sim \pi_\phi(\cdot\vert{s_t})$\;
\STATE Observe reward $r_{t+1}$ and new state $s_{t+1}$\;
\STATE Store ($s_t, a_t, r_{t+1}, s_{t+1}$) in replay buffer $\mathcal{D}$\;
\STATE if $s_{t+1}$ is terminal, reset environment state\;
\IF{the size of replay buffer $\mathcal{D} > m$}
    \FOR{gradient step $j$ in range (1, $N_G$), }
    \STATE Sample a batch from replay buffer $\mathcal{D}$ with size $m$ and calculate $\hat{\nabla}_{\theta_i} J_Q(\theta_i), \hat{\nabla}_\phi J_\pi(\phi)$ \;
    \STATE Update soft Q-function $\theta_i \leftarrow \theta_i - \lambda_Q \hat{\nabla}_{\theta_i} J_Q(\theta_i)$ for $i \in \{1, 2\}$\;
    \STATE Update policy network $\phi \leftarrow \phi - \lambda_\pi \hat{\nabla}_\phi J_\pi(\phi)$ \;
    \STATE Update target network $\bar{\theta}_i \leftarrow \tau \theta_i + (1-\tau)\bar{\theta}_i$ for $i \in \{1, 2\}$\;
    \STATE Adjust temperature $\alpha \leftarrow \alpha - \lambda \nabla_\alpha J(\alpha)$\;
    \ENDFOR
\ENDIF
 \ENDFOR
\end{algorithmic}
\end{algorithm}

\section{Experimental results}
\label{sec: experiment}

In this section, we conducted two categories of experiments: 1) to identify the optimal parameter settings for the proposed method, FreeMesh-RL and 2) to demonstrate the framework performance in scalability, generalizability, and mesh quality.

\subsection{Implementation settings}
This section examines the optimal settings of RL algorithm selection, training domain selection, SAC implementation, state representation, action space, and reward function design. All the experiments are conducted on a computer with an i7-8700 CPU and an Nvidia GTX 1080 Ti GPU with 32 GB of RAM. To evaluate the learning performance, we calculate the average return for ten evaluation episodes at every 10k time steps.

\subsubsection{RL method selection}
\begin{figure}
    \centering
    \includegraphics[width=\textwidth]{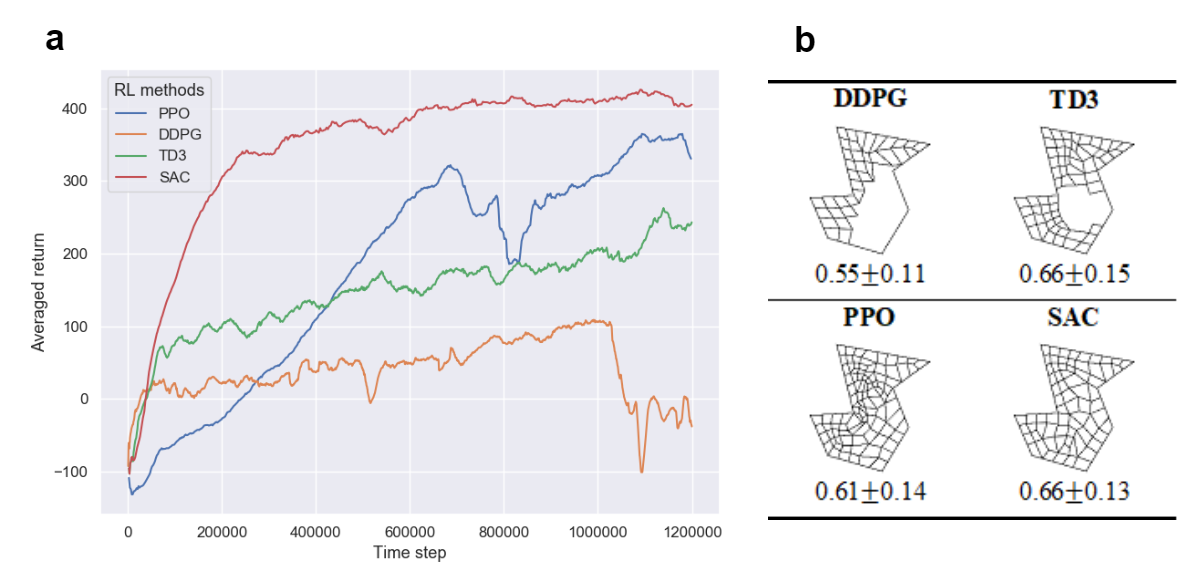}
    \caption{Learning efficiency and performance comparison of various RL algorithms. (a) Four types of RL algorithms, PPO, DDPG, TD3, and SAC, are chosen to compare the learning efficiency over the same training domain. (b) The performance of the policies learned by the four methods is examined on a domain. Their element quality in each mesh is computed. DDPG and TD3 cannot complete the mesh because of the early convergence of their meshing policy.}
    \label{fig:rl methods comparison}
\end{figure}

To be applied to real engineering problems, mesh generation requires a robust and stable policy to be learned without complicated hyperparameter tuning. We compared the learning efficiency of four types of RL algorithms (i.e., PPO, DDPG, TD3, and SAC) over the same training domain. With same hyperparameter settings in their original literature \citep{raffin2021stable}, the results are compared in Fig. \ref{fig:rl methods comparison} (a). In the early stage of training ($<$ 5e4), DDPG, TD3, and SAC achieve faster learning performance than the PPO method. Then the learning speeds of DDPG and TD3 decrease, while SAC maintains high speed and converges to a stable meshing policy. Although PPO gradually achieves suboptimal performance, its policy severely oscillates. 

We also tested all the learned policies on a domain, as illustrated in Fig. \ref{fig:rl methods comparison} (b). The policies learned by DDPG and TD3 do not converge and cannot successfully mesh the domain. PPO can easily generate irregular elements, causing poor mesh quality, whereas SAC has successfully meshed the domain. Without further hyperparameter adjustment, the SAC method achieves the fastest learning performance and the most stable meshing policy. SAC appears to be an optimal method for mesh generation.

\subsubsection{Training domain selection}
In a real engineering environment, the meshing domains have diverse shapes and topological structures. The basic domain features include sharp angles, bottleneck regions, and unevenly distributed boundary segments for a single connected 2D geometry. Therefore, the training domain should match those criteria to ensure the richness of the samples. The meshing boundary changes constantly during element generation, as shown in Fig. \ref{fig:rl mesh}. The total number of intermediate boundaries is equivalent to the number of elements generated. This process will also increase the sample diversity.

\begin{figure}
    \centering
    \includegraphics[width=\textwidth]{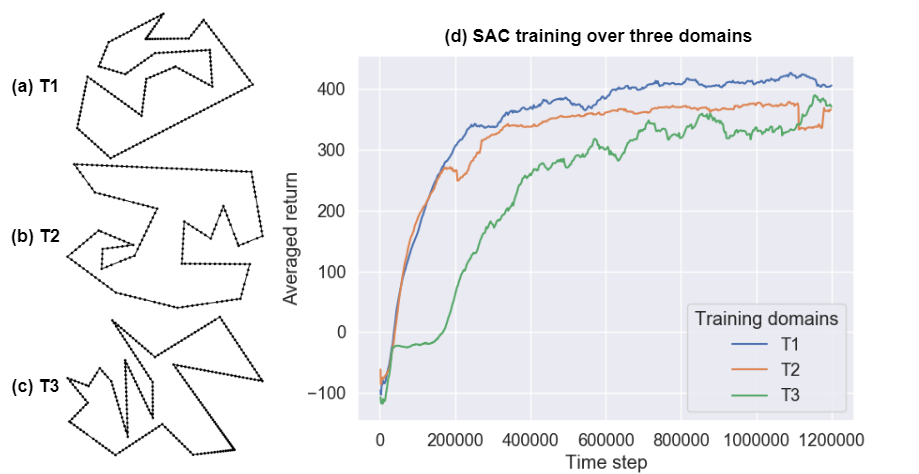}
    \caption{Learning difficulty comparison over different training domains. (a)-(c) Three types of training domains, T1, T2, and T3, are designed based on the identified criteria. (d) The training results over three domains using SAC. Domain T1 achieves a fast and stable learning efficiency which has an easier meshing policy to learn than the other two domains, whereas domain T3 is the hardest one to converge to a stable policy because multiple sharp angles exist.}
    \label{fig:training domain comparison}
\end{figure}

To find the appropriate training domain, we designed three candidate domains, T1, T2, and T3, based on the above criteria, as shown in Fig. \ref{fig:training domain comparison} (a-c). The training results using SAC are shown in Fig. \ref{fig:training domain comparison}(d). The agents in both domains T1 and T2 achieve fast learning speed and converge to a stable meshing policy, but T1 has less learning fluctuation. The agent spends more learning time in domain T3 because of the difficulties of having more sharp angles and bottlenecks. We choose domain T1 as the training domain because of its optimal steadiness and learning efficiency. 
As shown in the rest of experiments, the model trained with T1 has strong generalizability.

\subsubsection{SAC implementation}
\begin{figure}
    \centering
    \includegraphics[width=\textwidth]{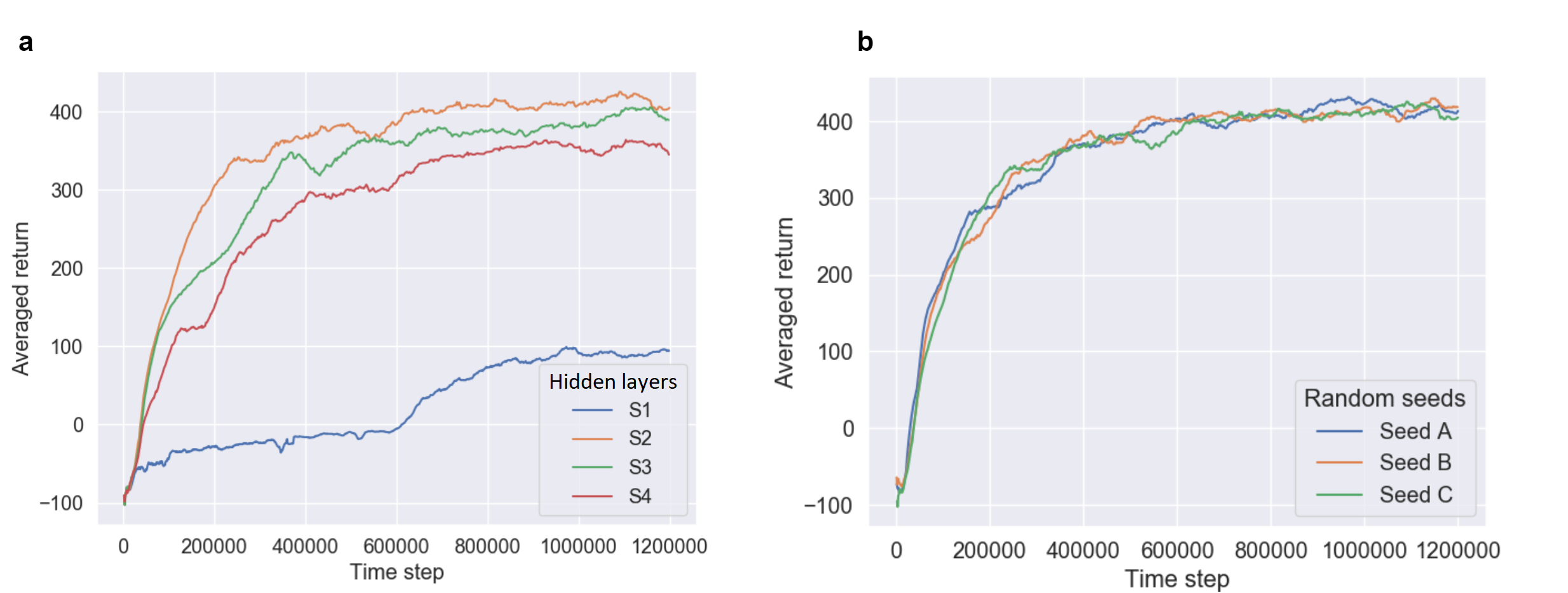}
    \caption{Comparisons of hidden layer configurations and random seeds for the SAC algorithm. (a) Comparison of four types of hidden layer configurations: S1-S4 represent four different neural network structures in the hidden layers, including [32, 32], [128, 128, 128], [64, 64, 64, 64, 64], and [32, 128, 128, 128, 64, 32], respectively. Configuration S2 achieves a fast and stable learning result. (b) Random seed comparison: three different random seeds, A (356), B (567), and C (999), are chosen to examine the learning difference. Certainly, SAC is not sensitive to those random seeds and could achieve stable learning results.}
    \label{fig:nn structures comparison}
\end{figure}

To identify the optimal NN hidden layer setting (i.e., the number of hidden layers and neurons in each layer) in SAC, we compared four different configurations, S1-S4, from the perspective of learning efficiency. The results are compared in Fig. \ref{fig:nn structures comparison} (a). Configuration S1 has the poorest performance in learning the policy. The network structure S2 with three hidden layers, [128, 128, 128], achieves the best learning performance while having fewer parameters than S3 and S4. It is used in the article to approximate the soft Q-function, policy, and target networks.

A random seed is a common parameter in RL, which often affects learning performance. We selected three random seeds and tested their impact on policy learning, as shown in Fig. \ref{fig:nn structures comparison} (b). It turns out that the SAC learning process is not sensitive to those seeds, which is one of its advantages compared with other RL methods. The other hyperparameters used for SAC are listed in Table \ref{tab:hyperparameters}.

\begin{table}[h!]
    \centering
     \caption{Training hyperparameters for the SAC algorithm.}
    \label{tab:hyperparameters}
    \begin{tabular}{ ccc }
   \toprule[1pt]
    Parameter & Description & Value \\ \hline
     $N_\mathcal{D}$ &  Experience pool size & 1e6 \\
    $m$ &  Minibatch size & 256 \\
    $\gamma$ &  Discount factor & 0.99 \\
    $\lambda_Q, \lambda_\pi$ & Learning rate & 3e-4 \\
    $N_T$ & Total time steps & 1.2e6 \\
    $N_G$ & Gradient steps & 1 \\
    $\tau$ & Soft update factor & 5e-3\\\bottomrule[1pt]
\end{tabular}
\begin{tablenotes}
      \small
      \item 
    \end{tablenotes}
\end{table}

\subsubsection{Agent's view of environmental state}
\begin{figure}
    \centering
    \includegraphics[width=0.8\textwidth]{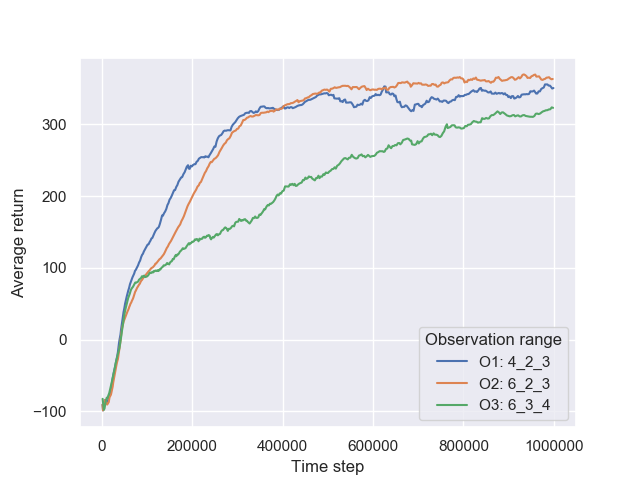}
    \caption{Learning efficiency comparison of different agent's observation ranges. The observation range is formed by $\beta\_n\_g$, which represents the radius of the fan shape in the state, the number of neighboring vertices on the left and right sides of the reference vertex, and the number of vertices in the fan-shaped area. This range determines how far and how much information the agent will observe in the meshing environment.}
    \label{fig:state comparison}
\end{figure}

The state is the agent's observation of the environment and provides a decision basis for the agent. As partial observation is adopted in this method, it is necessary to decide the observation range for the agent to learn the meshing policy effectively. The range of the observation is controlled by three parameters, $\beta$ (in Equation \ref{eqn: fan radius}), $n$ and $g$ (in Equation \ref{eqn: state function}). The parameter $\beta$ controls how far in the fan-shaped area the agent can observe from the selected reference vertex while the other two parameters determine how many vertices the agent perceives around the reference vertex. The learning performance is compared with three types of settings, $O1$ ($\beta=4, n=2, g=3$), $O2$ ($\beta=6, n=2, g=3$), and $O3$ ($\beta=6, n=3, g=4$). The results are compared in Fig. \ref{fig:state comparison}. By comparing $O1$ and $O2$, it can be found that further observation contributes to more return. This is because the agent could adjust the position of the candidate vertex in advance to avoid a conflict with the remaining boundary. On the other hand, the more vertices in the fan-shaped areas are considered, the more information will be embedded in the state. Therefore, increasing $g$ could slow the learning speed and delay policy convergence. This phenomenon can be observed when comparing $O2$ and $O3$. The more information the agent observes, the more time is needed to build the correlation. 

We also evaluated the influence of the observation ranges on the meshing performance from the aspects of element quality ($\eta^e$ in Equation \ref{element quality}), the number of elements generated, and time cost. Table \ref{tbl:meshing_observation_comaprison} presents those performance comparisons. The policies learned by the observations of $O1$ and $O2$ have a small difference in element quality, whereas the one learned by $O3$ is relatively poorer because of the slow learning speed. With the increase in observed information, the number of generated elements also increases. This is reasonable because the agent could make prudent decisions to avoid collision with the sensed boundary in a fan shape when it observes farther and broader. As illustrated by the sample meshes, the interior area has smaller and more regular elements by $O2$ and $O3$ as the boundary is updated inwardly. The meshing speed hinges on the number of generated elements, which is indirectly influenced by the observation ranges because the agent tends to make smaller elements if it observes a potential collision with the sensed boundary. Ideally, the more information the agent observes, the more accurate the decision it can make and the more learning time it requires to converge. To achieve a trade-off between the computational cost and mesh quality, the observation range $O2$ is used to represent the state in this article. 

\begin{table*}[h!]
  \centering
\begin{adjustbox}{width=\textwidth}
\begin{threeparttable}
\caption{Meshing results comparison of different agents' observation ranges. The observation ranges are $O1$, $O2$, and $O3$. The performance is evaluated from three indicators, including element quality, the number of generated elements, and time cost. A sample mesh (without smoothing) for each range is also represented. All the experiments are repeated ten times over the same domain for each observation.}
\label{tbl:meshing_observation_comaprison}
  \begin{tabular}{ c c c c c }
    \toprule[1pt]
    Observation & Sample mesh & Element quality & \#elements & Time cost (s) \\ \hline
      $O1$: $4\_2\_3$
    &
      \begin{minipage}{.3\textwidth}
      \includegraphics[width=\linewidth]{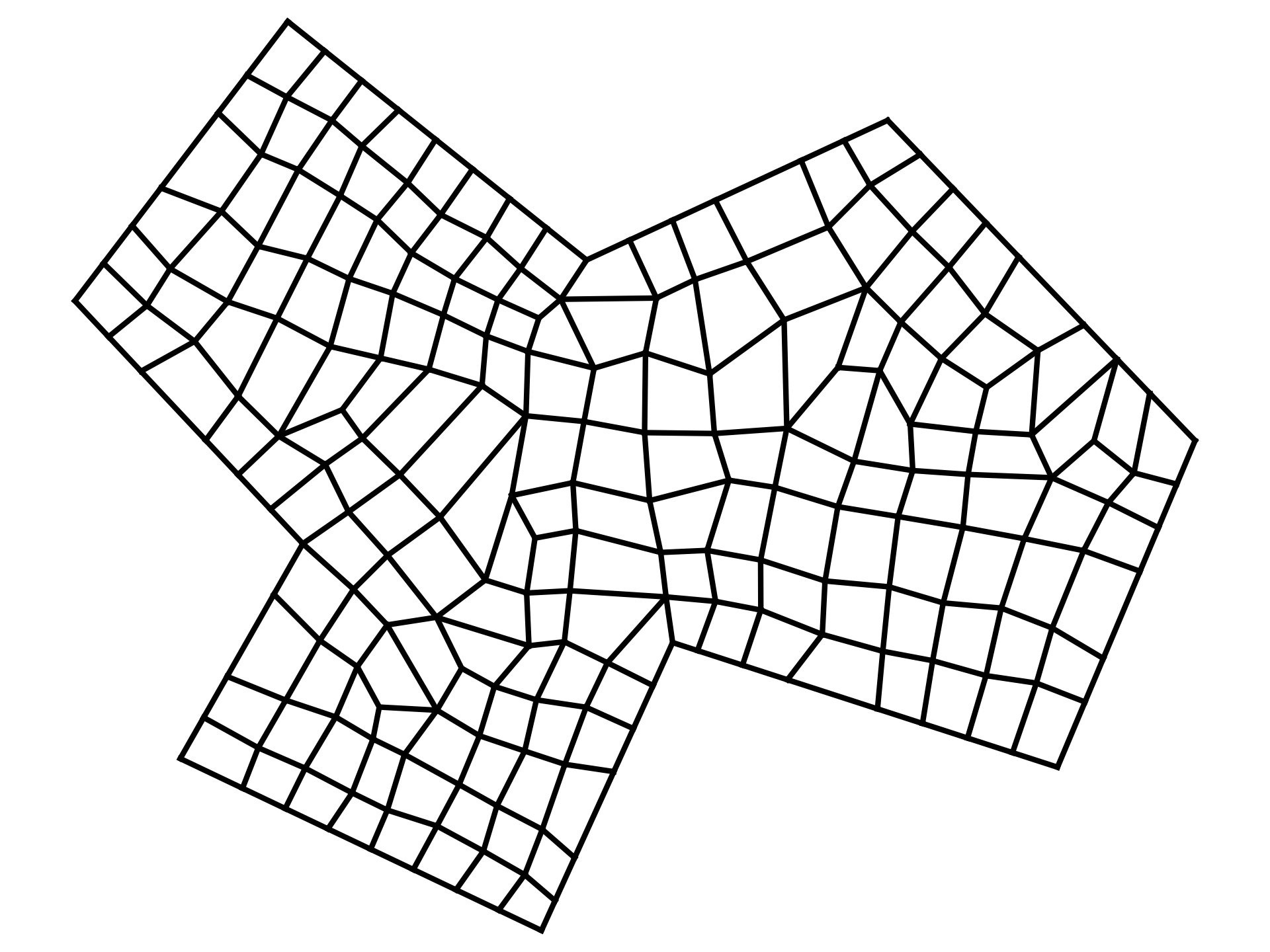}
    \end{minipage}
    & 
    0.722 $\pm$ 0.14
    &
    198.8 $\pm$ 22.11
    & 
    0.6 $\pm$ 0.09
    \\ 
    $O2$: $6\_2\_3$
    &
      \begin{minipage}{.3\textwidth}
      \includegraphics[width=\linewidth]{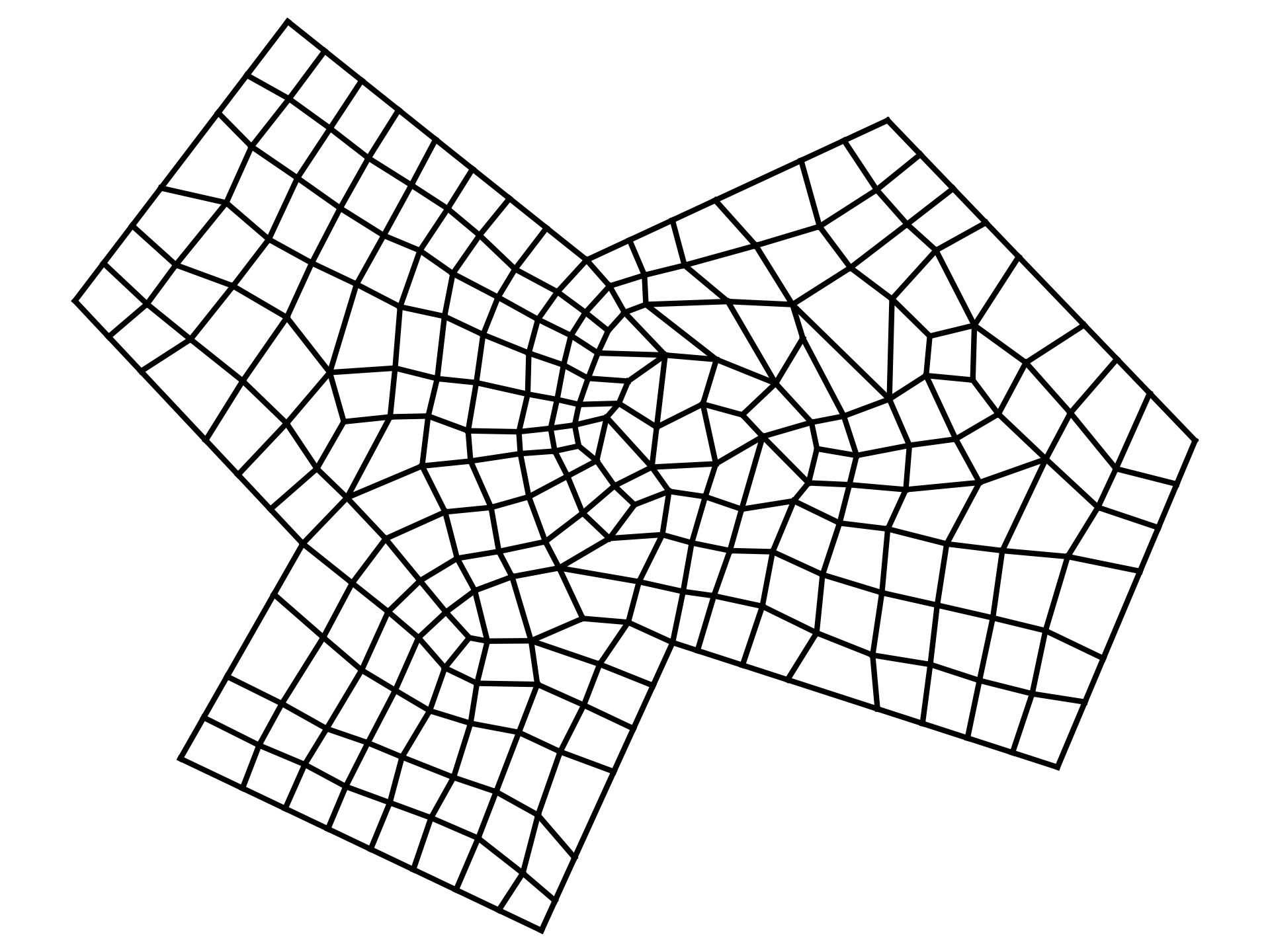}
    \end{minipage}
    & 
    0.712 $\pm$ 0.13
    &
    212.8 $\pm$ 24
    & 
    0.64 $\pm$ 0.08 \\
    $O3$: $6\_3\_4$
    &
      \begin{minipage}{.3\textwidth}
      \includegraphics[width=\linewidth]{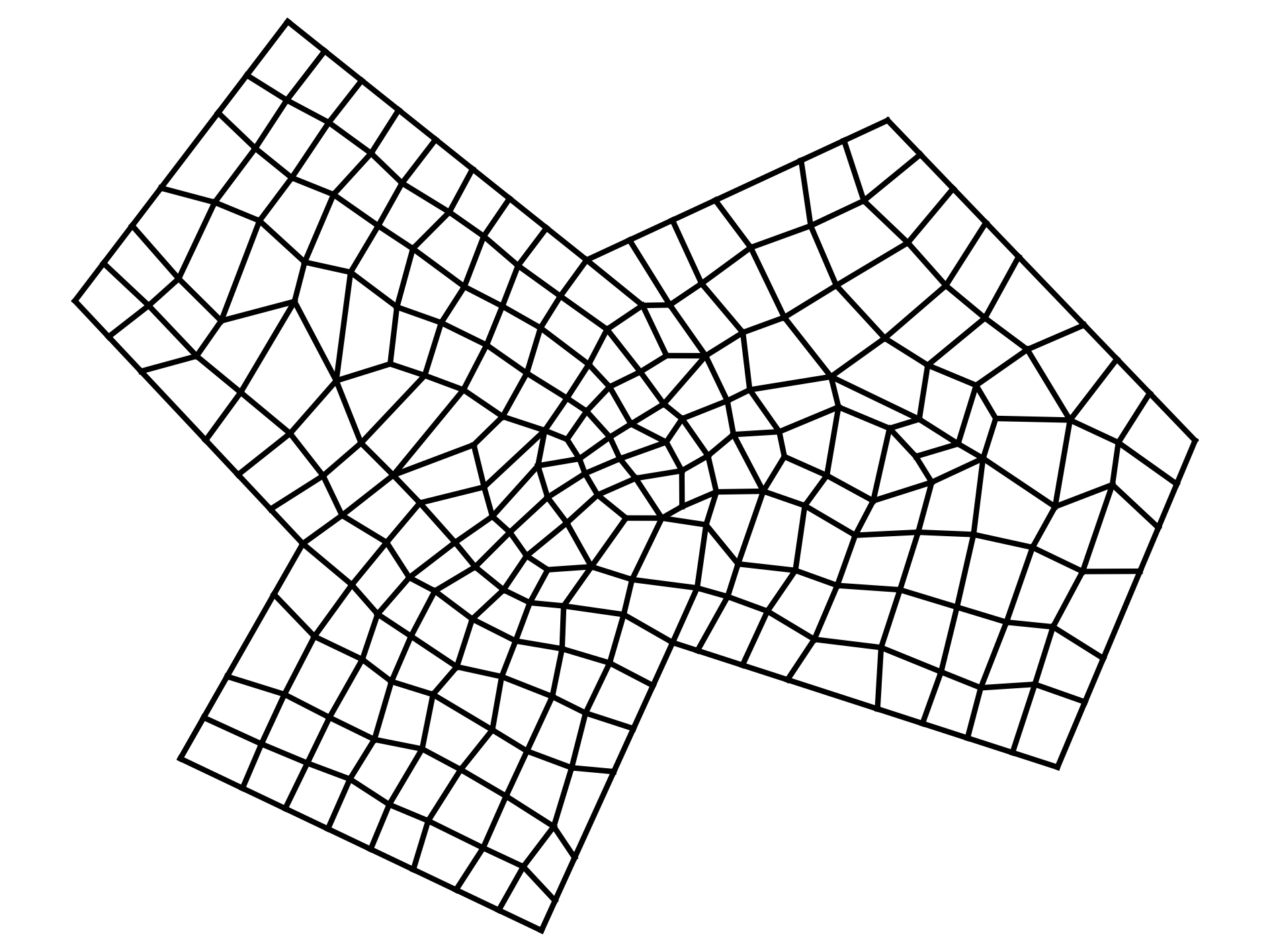}
    \end{minipage}
    & 
    0.689 $\pm$ 0.137
    &
    217.1 $\pm$ 11.16
    & 
    0.7 $\pm$ 0.13
    \\ \bottomrule[1pt]
  \end{tabular}
\begin{tablenotes}
      \small
      \item \#elements - the number of generated elements.
    \end{tablenotes}
\end{threeparttable}
\end{adjustbox}
\end{table*}

\subsubsection{Action space}
The agent's action space defines the shape and size scale of each element to be generated, which influences the mesh quality and policy's learning efficiency. This section will examine the appropriate size of the search space, i.e., the coordinate space to select the candidate vertex in forming a quadrilateral element. The coordinate space, specified by the radius in Equation \ref{eqn: radius}, is defined by weight $\alpha$ and a base length $L$. The number of vertices in calculating the base length is equivalent to the number of neighboring vertices of the reference vertex in the state (see the parameter $n$ in Equation \ref{eqn: state function}). The ideal observation range (i.e., state) is determined as $O2$ in the previous section. The parameter $n$ is hence set to 2.

\begin{figure}
    \centering
    \includegraphics[width=0.8\textwidth]{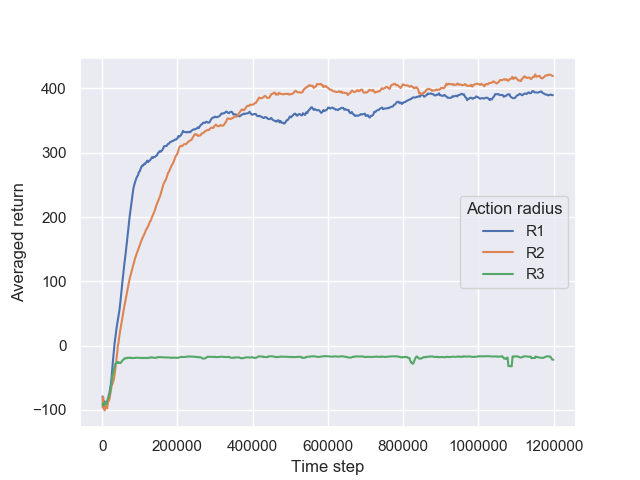}
    \caption{Learning efficiency comparisons of three types of radius settings in the action. The appropriate radius is necessary for optimal mesh quality and learning efficiency. These settings are $R1$ ($\alpha = 1$), $R2$ ($\alpha = 2$), and $R3$ ($\alpha = 3$). A larger $\alpha$ indicates a larger coordinate space for searching a candidate vertex used to form a quadrilateral element.}
    \label{fig: action radius}
\end{figure}

The size of the search space is defined by the radius of the viewing fan shape, determined by the weight $\alpha$. To find the appropriate radius, three types of settings, $R1$ ($\alpha = 1$), $R2$ ($\alpha = 2$), and $R3$ ($\alpha = 3$), are examined during meshing policy learning over the same domain (see Fig. \ref{fig:training domain comparison} (a)). Their learning results are shown in Fig. \ref{fig: action radius}. The radius $R1$ achieves the fastest learning speed and converges earlier than the other two settings. Because its search space is the smallest, action exploration is less needed. Although the radius $R2$ is slightly slower in convergence than $R1$, it facilitates the highest reward return indicating optimal mesh quality, due to the larger search space. Unfortunately, the agent fails to learn an efficient policy by radius $R3$. The large search space contains sparse efficient actions, and makes agent's exploration harder. Therefore, the radius $R2$ is used in the present article. 

\subsubsection{Reward function}
There are three terms in the reward function: element quality $\eta^e_t$, the quality of the remaining boundary $\eta^b_t$, and density $\mu_t$. The first two terms guarantee the element quality and the ease of continuous meshing. The last term controls the meshing density by the parameter $\upsilon$, which adjusts the minimum element size tolerated, and ensures the mesh termination within finite steps. The first two terms are necessary and not changeable once the quality requirement is determined, whereas the third term can be adjusted according to different requirements for mesh density.

\begin{figure}
    \centering
    \includegraphics[width=0.8\textwidth]{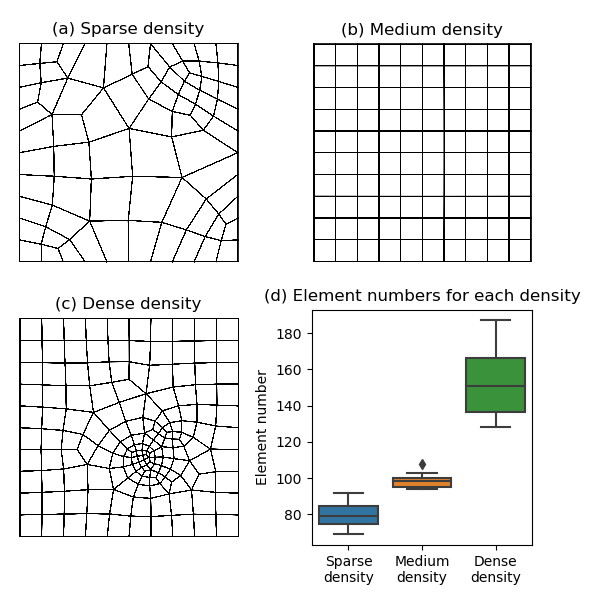}
    \caption{Different mesh densities controlled by the reward function. Three different densities from sparse to dense, (a)-(c), are controlled by the parameters $\upsilon = 1.5, \upsilon = 1$, and $\upsilon = 0.5$, respectively. Subfigure (d) shows the results of the number of generated elements by three types of densities averaged over ten episodes.}
    \label{fig:mesh density}
\end{figure}

To achieve the optimal mesh density, three different parameters are compared, including sparse ($\upsilon = 1.5$), medium ($\upsilon = 1$), and dense ($\upsilon = 0.5$) settings. The results are compared in Fig. \ref{fig:mesh density} (a)-(c). We also examine the number of elements generated by each density, as shown in Fig. \ref{fig:mesh density} (d). The results are averaged over 10 episodes. The difference in the number of elements between sparse and medium is approximately 20, while the difference between medium and dense density is approximately 50 elements in the testing domain. The medium density is ideal and used across all the remaining experiments.

\subsection{Effectiveness evaluation}
The effectiveness of FreeMesh-RL is evaluated from the perspectives of scalability, generalizability, and mesh quality. The model used in the following experiments is trained on the same domain (i.e., domain T1 in Fig. \ref{fig:training domain comparison} (a)) without any adjustment.

\subsubsection{Scalability verification}
To validate the scalability of the learned meshing policy, we constructed three geometry domains with the same shape but different vertex densities (i.e., 6.8: 9.9: 20.1, as shown in Table \ref{tab:scalability_comparison}) on the boundaries. Three domains are meshed by the same RL model, and the results are shown in Table \ref{tbl:density_comaprison}. It can be seen that all the domains have successfully meshed; the elements on the boundaries are denser than in the interior area, which is beneficial for reducing computational burden; and the transitions between dense and coarse meshes are smooth. The meshing speed over three domains is approximately 237 elements per second on average. Consequently, the results show that the learned mesh policy achieves good scalability to different scales of the meshing problem and is not constrained to the 
density requirements
of the training domain. 
This scalability is a special manifestation of the broader generalizability to be discussed next.

\begin{table*}[h!]
  \centering
  \caption{Meshing the same domain with different boundary segment densities by FreeMesh-RL. The boundary shapes of domains 1-3 are the same, but the number of vertices on the boundary is from low to high. The scalability is examined by meshing all the domains with the same trained model.}
  \begin{adjustbox}{width=\textwidth}
\begin{threeparttable}
  \label{tbl:density_comaprison}
  \begin{tabular}{ ccc }
     \toprule[1pt]
    Domain 1 & Domain 2 & Domain 3 \\ \hline
    \begin{minipage}{.33\textwidth}
      \includegraphics[width=\linewidth]{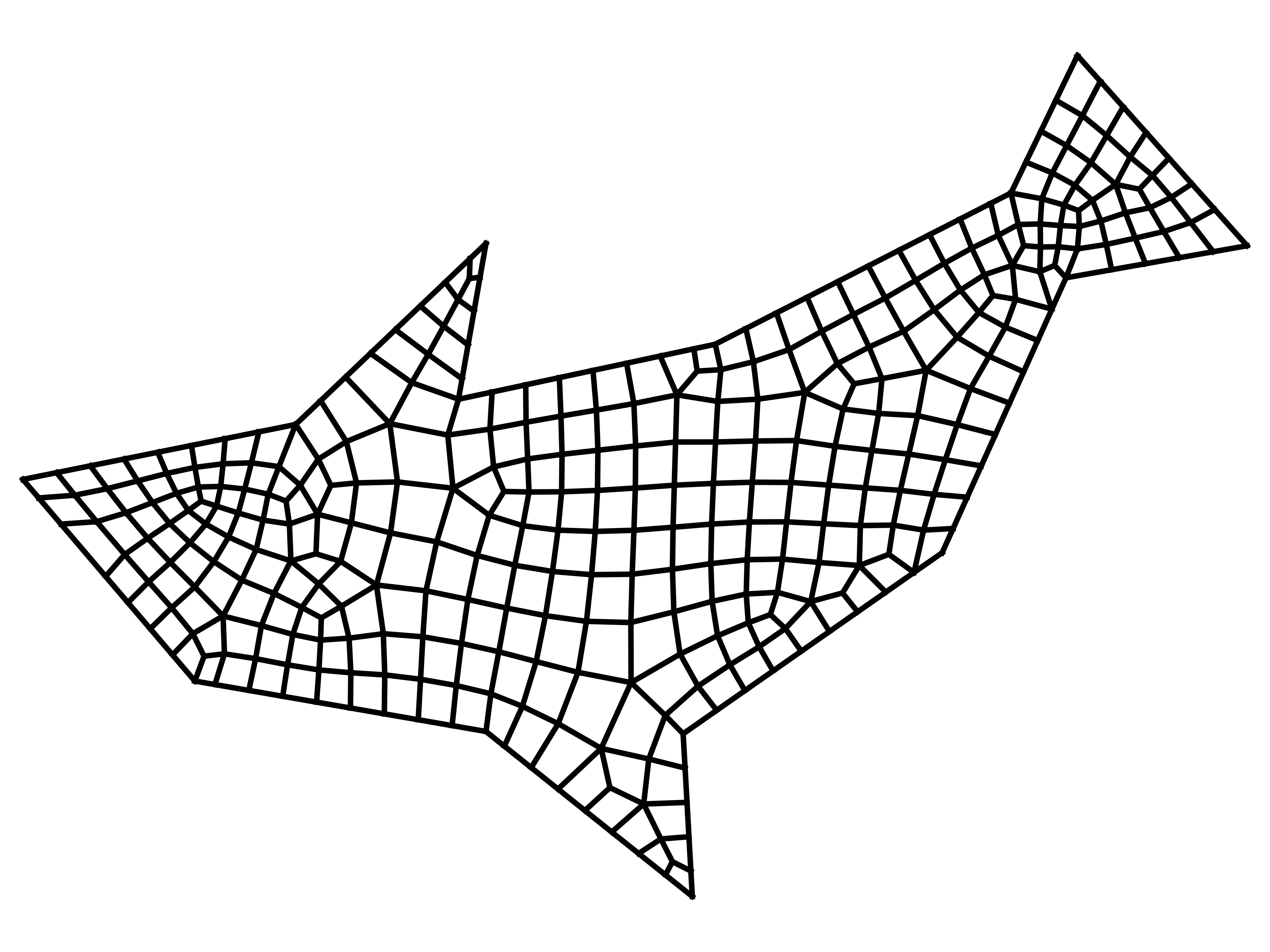}
    \end{minipage}
    &
      \begin{minipage}{.33\textwidth}
      \includegraphics[width=\linewidth]{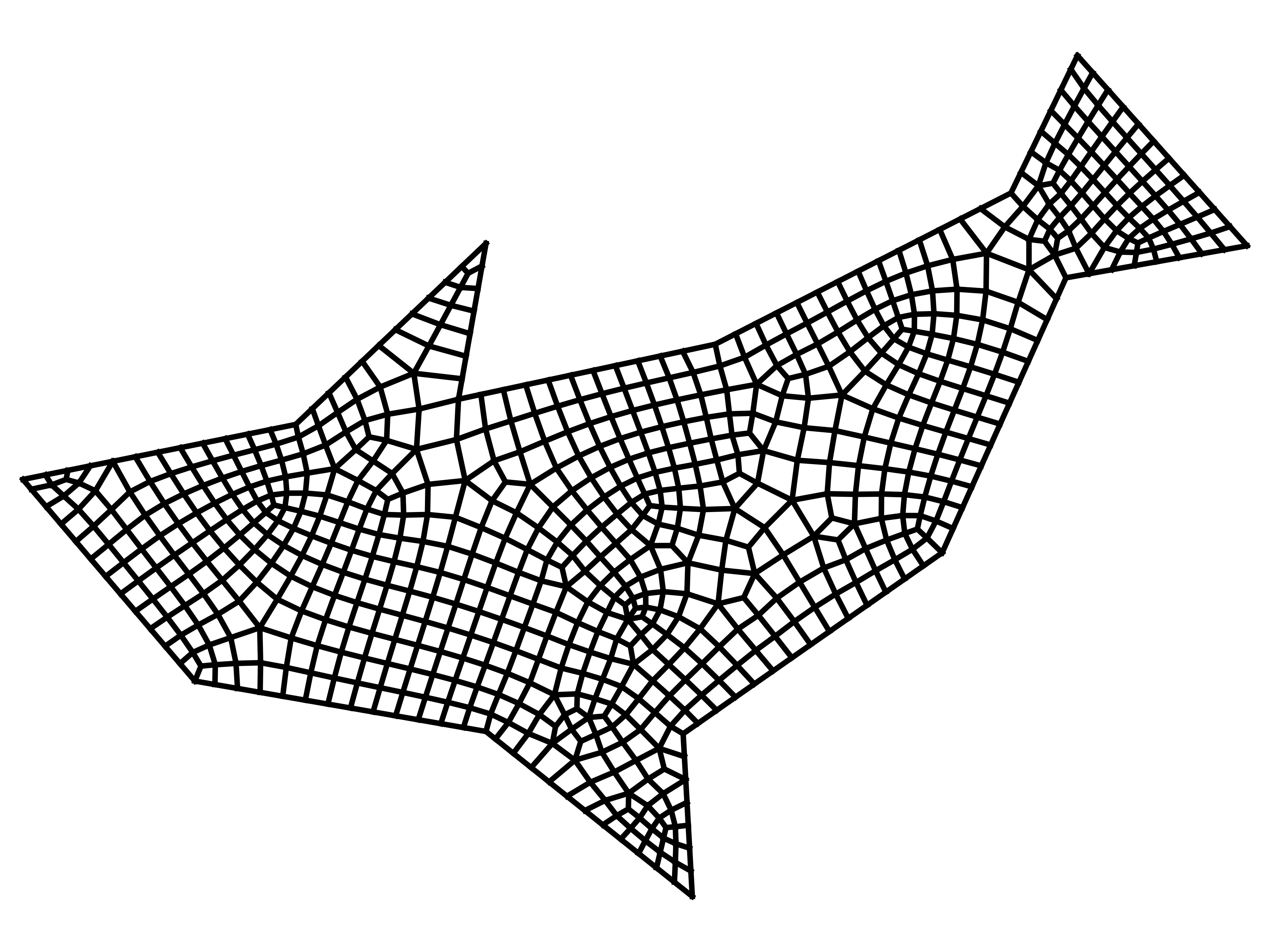}
    \end{minipage}
    & 
      \begin{minipage}{.33\textwidth}
      \includegraphics[width=\linewidth]{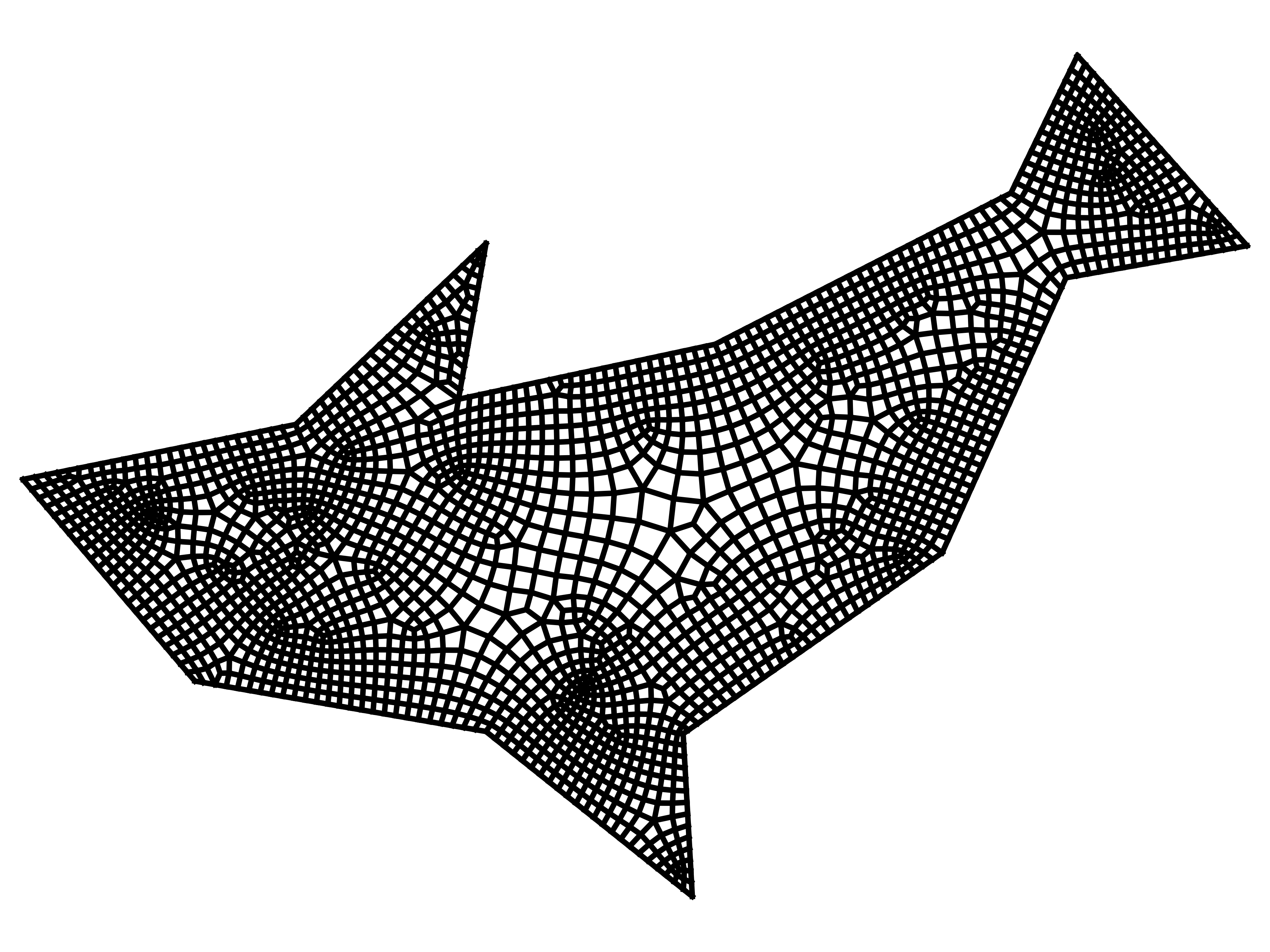}
    \end{minipage}
    \\ \bottomrule[1pt]
  \end{tabular}
  \begin{tablenotes}
      \small
      \item 
    \end{tablenotes}
    \end{threeparttable}
    \end{adjustbox}
\end{table*}

\begin{table}[h!]
    \centering
     \caption{Geometrical details of three domain boundaries and their generated meshes. The perimeter, the number of vertices and vertices per unit length on the initial boundaries of three domains are compared as well as their number of generated elements and meshing time.}
    \label{tab:scalability_comparison}
    \begin{tabular}{cccc}
   \toprule[1pt]
    & Domain 1 & Domain 2 & Domain 3 \\ \hline
   \#vertices  &  102 & 150 & 304 \\
   Perimeter  &  15.1 & 15.1 & 15.1 \\
   \#vertices per unit length &  6.8 & 9.9 & 20.1 \\
   \#elements & 289 & 665 & 2157 \\
   Execution time (s) & 0.9 & 2.6 & 15.3 \\\bottomrule[1pt]
\end{tabular}
\begin{tablenotes}
      \small
      \item \#vertices - the number of vertices on the boundary.
      \item \#elements - the number of generated elements.
    \end{tablenotes}
\end{table}

\subsubsection{Generalizability verification}
\begin{figure}
    \centering
    \includegraphics[width=\textwidth]{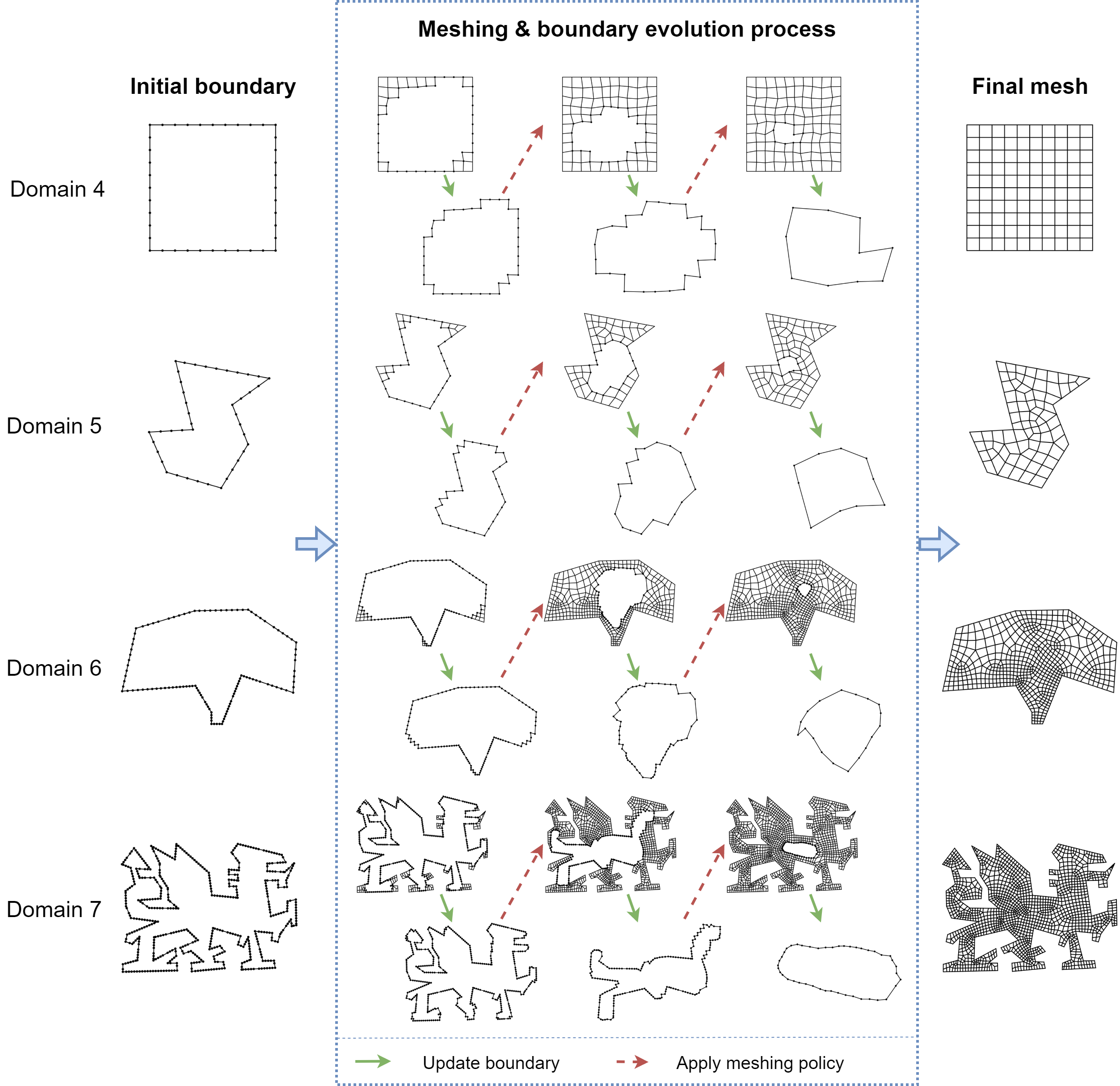}
    \caption{Model generalizability verification. Four different domains with shapes from simple to complex are selected; their meshing procedures are presented to exhibit the boundary evolution process and general meshing pattern by the learned policy. No matter how simple an initial shape was, the intermediate shapes can be complex; no matter how complex the shape of the initial boundary was, it gradually evolves into a smooth oval shape in the center of the domain.}
    \label{fig:domain generic}
\end{figure}

To verify the generalizability of the obtained meshing model, we meshed four different domains, domains 4-7, using the same model, trained with domain T1 as shown in Fig. \ref{fig:training domain comparison}. The shapes of the domains are from simple to complex. The meshing process and results are shown in Fig. \ref{fig:domain generic}. The general meshing process starts from the boundary and advances inwardly to the central area of the domain. Regardless of the complexity of the initial domains, the boundary will evolve into various intermediate shapes with the meshing process guided by the learned meshing policy. The dynamic changes of the domain boundary also reflects the generalizability, i.e., how good the policy to handle different shapes.

\begin{figure}
    \centering
    \includegraphics[width=0.8\textwidth]{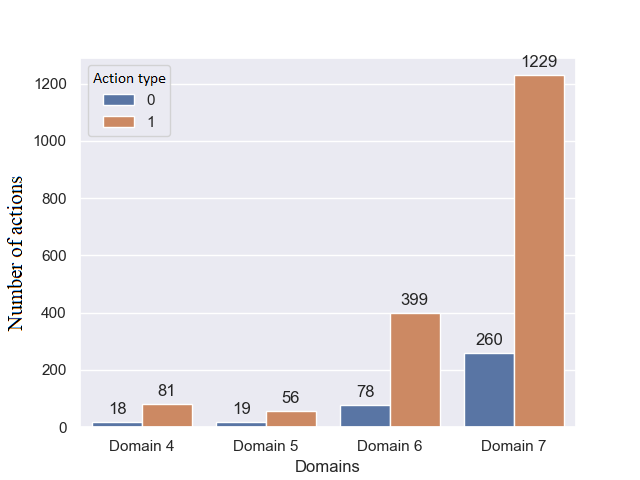}
    \caption{The execution times of the meshing policy over the four domains. The meshing policy mainly consists of two types of actions: types 0 and 1. The number over each bar indicates the execution number of each action type.}
    \label{fig: policy execution}
\end{figure}

The number of actions executed in meshing each of the four domains is shown in Fig. \ref{fig: policy execution}. The meshing policy mainly consists of two types of actions (see type 0 and type 1 in Fig. \ref{fig:action space}), and the execution number of type 1 is approximately $80\%$ of the total number. Each  action, a single execution of the policy, will produce an element. The number of intermediate boundaries equals the number of elements generated. For example, in domain 7, there are 1489 (the sum of execution times of type 0 and type 1) intermediate boundaries in total. The meshing policy succeeds in meshing 1489 different domain boundaries. It can be noted that all the intermediate boundaries of the four domains gradually become an oval shape in Fig. \ref{fig:domain generic}. This demonstrates that the meshing policy can solve complex boundary situations with sharp angles and create smooth and easy-to-handle situations for future element generation. 

Although the geometry shape can be diverse and infinite in real engineering applications, our state representation with a partial boundary and the use of a reference vertex as the origin make our model general and able to handle various boundary shapes. As demonstrated above, the obtained policy is able to mesh hard situations while maintaining the high quality of the remaining boundaries. Therefore, the most challenging part of the whole meshing process is to mesh the initial boundary situations of a given domain. However, it can be easily solved by providing training domains with sufficient difficulties (in Fig. \ref{fig:training domain comparison} (a-c)). To conclude, the proposed method can be applied to arbitrary domains and is general to various geometries.

\subsubsection{Comparison to conventional methods}
To evaluate the meshing performance, we compare the quality of meshes by FreeMesh-RL with two other representative meshing approaches over three predefined 2D domains (i.e., domains D7-9). These domains possess different features, including sharp angles, bottleneck regions, unevenly distributed edges, and holes, to increase the testing diversity. The two conventional meshing approaches are Blossom-Quad \citep{remacle_blossom-quad_2012} and Pave \citep{blacker_paving_1991, white_redesign_1997}. Blossom-Quad is an indirect method that generates quadrilateral elements by finding the perfect matching of a pair of triangles generated in advance. The method is implemented by an open source generator Gmsh \citep{geuzaine_gmsh_2009}. Pave is another state-of-the-art meshing method for directly generating quadrilateral elements, implemented by the CUBIT software \citep{blacker_cubit_2016}.

The meshing results are shown in Table \ref{tbl:meshing_results_comaprison}. Although all the methods can complete the meshes for the three domains, there are some subtle differences. Only FreeMesh-RL generates fully quadrilateral meshes. The other two methods have difficulties in discretizing the domains into full quadrilaterals, containing triangles in domains (marked in yellow color in each domain). Specifically, Blossom-Quad has a problem in handling domains with sharp angles along the boundary, while Pave has issues in the interior of the domain. Extra operations (e.g., clean-ups) are thus usually required to eliminate those triangles or bad elements. Another advantage of FreeMesh-RL is that the generated mesh can smoothly transition from very small to large elements over a short distance, as shown in Table \ref{tbl:meshing_results_comaprison} (domain 9), which is beneficial in reducing the computational burden during simulations \citep{shewchuk_unstructured_2012}. 

\begin{table*}[h!]
  \centering
\begin{adjustbox}{width=\textwidth}
\begin{threeparttable}
\caption{Meshing results comparison. Three domains are designed to examine the meshing performance based on the four identified criteria: a. sharp angle, b. narrow region, c. unevenly distributed boundary segments, and d. having a hole inside. Both domains 7 and 8 possess features a and b, whereas domain 9 has features c and d. Two representative methods, Blossom-Quad and Pave, are chosen to compare the meshing performance with the proposed method, FreeMesh-RL.}
\label{tbl:meshing_results_comaprison}
  \begin{tabular}{ c c c c }
    \hline
    Algorithms & Domain 7 & Domain 8 & Domain 9 \\ \hline
    \begin{minipage}{.1\textwidth}
      Blossom-Quad
    \end{minipage}
    &
      \begin{minipage}{.3\textwidth}
      \includegraphics[width=\linewidth]{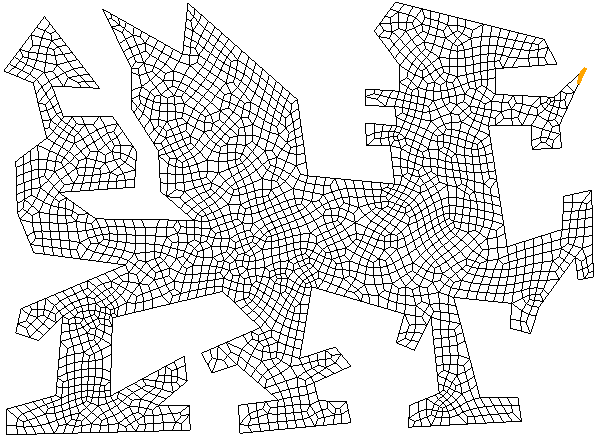}
    \end{minipage}
    & 
      \begin{minipage}{.3\textwidth}
      \includegraphics[width=\linewidth]{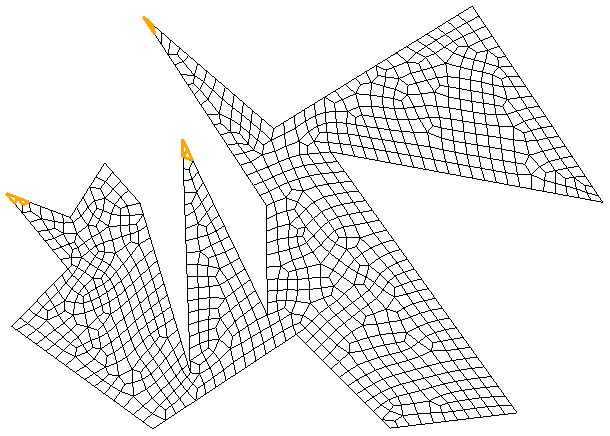}
    \end{minipage}
    &
    \begin{minipage}{.3\textwidth}
      \includegraphics[width=\linewidth]{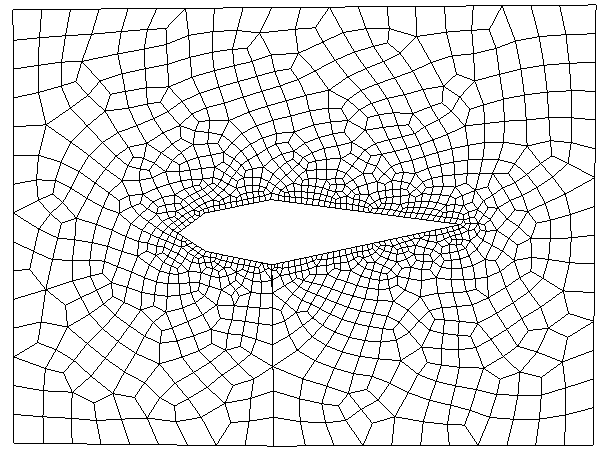}
    \end{minipage}
    \\ 
     \begin{minipage}{.1\textwidth}
      Pave
    \end{minipage}
    &
      \begin{minipage}{.3\textwidth}
      \includegraphics[width=\linewidth]{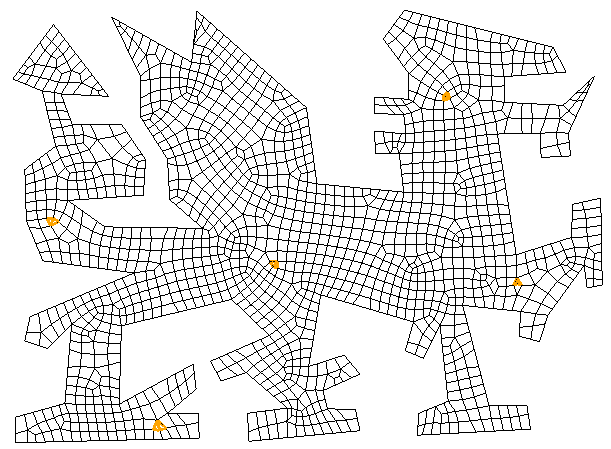}
    \end{minipage}
    & 
      \begin{minipage}{.3\textwidth}
      \includegraphics[width=\linewidth]{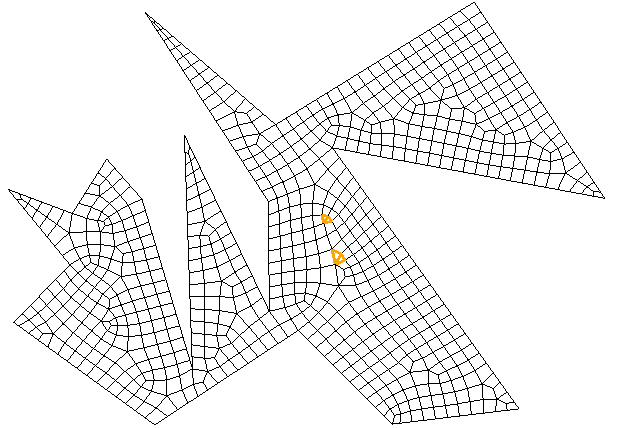}
    \end{minipage}
    &
    \begin{minipage}{.3\textwidth}
      \includegraphics[width=\linewidth]{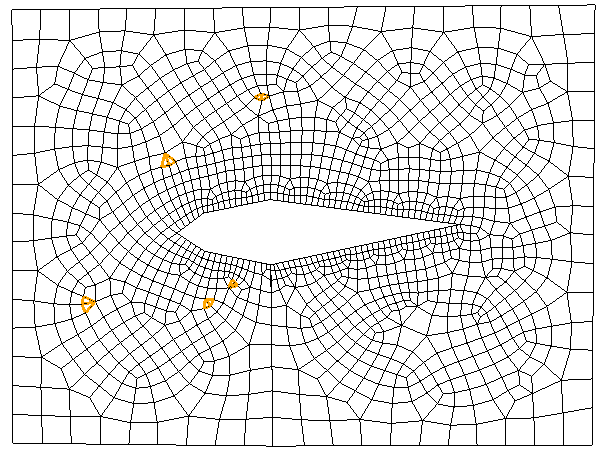}
    \end{minipage}
    \\ 
     \begin{minipage}{.1\textwidth}
      FreeMesh-RL
    \end{minipage}
    &
      \begin{minipage}{.3\textwidth}
      \includegraphics[width=\linewidth]{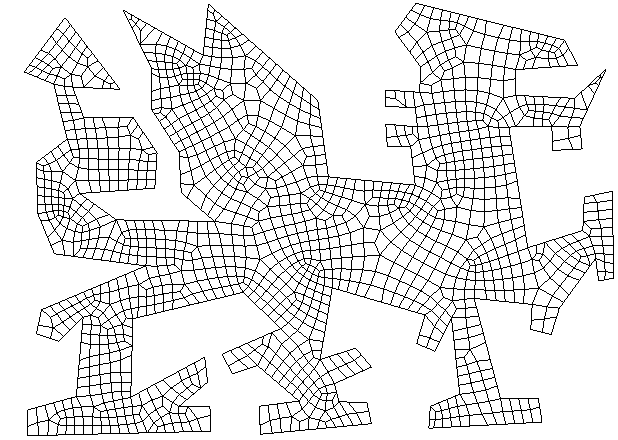}
    \end{minipage}
    & 
      \begin{minipage}{.3\textwidth}
      \includegraphics[width=\linewidth]{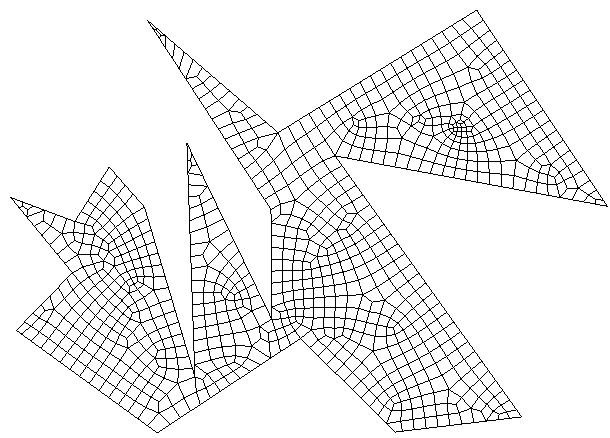}
    \end{minipage}
    &
    \begin{minipage}{.3\textwidth}
      \includegraphics[width=\linewidth]{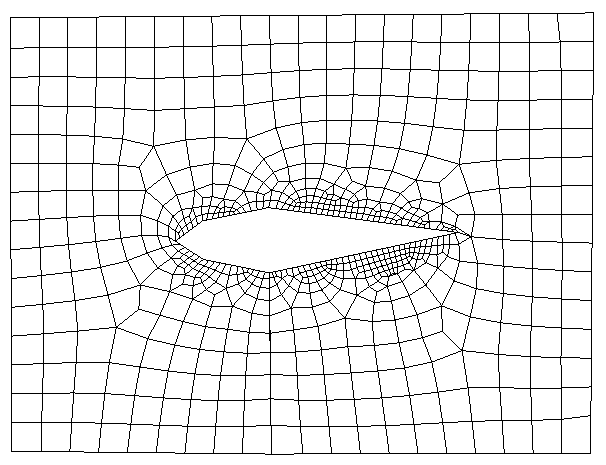}
    \end{minipage}
    \\ \hline
  \end{tabular}
\begin{tablenotes}
      \small
      \item The elements in yellow represent existing triangles in the meshes.
    \end{tablenotes}
\end{threeparttable}
\end{adjustbox}
\end{table*}

To quantitatively analyze the meshing results of the three methods, we selected eight common quality metrics, including 1) singularity, the number of irregular nodes whose number of incident edges in the interior of a mesh is not equal to four; 2) element quality, $\eta^e$ in Equation \ref{element quality}; 3) $\vert MinAngle-90^{\circ}\vert$; 4) $\vert MaxAngle-90^{\circ} \vert$; 5) scaled Jacobian, the minimum Jacobian \citep{knupp2000achieving} at each corner of an element divided by the lengths of the two edge vectors; 6) stretch, the degree of deformation; 7) taper, the maximum absolute difference between the value one and the ratio of two triangles’ areas are separated by a diagonal within a quadrilateral element; and 8) the number of triangles (\#triangles) \citep{pan_huang_wang_cheng_zeng_2021, knupp2006verdict}. Smaller singularity and taper mean better regularity, which can provide more accurate results for numerical simulations. A larger scaled Jacobian indicates higher convexity of the mesh, containing less inverted and flat quadrilateral elements. The existence of triangles indicates that the domain is not fully meshed by quadrilaterals and usually require extra treatments to eliminate.

All the measurement results are averaged over three domains and are shown in Table \ref{tab:averagaed_mesh_quality}. The proposed method, FreeMesh-RL, outperforms other methods in the indices of singularity, taper, scaled Jacobian, and number of triangles, while being comparable with other best performing indices by Pave. Pave achieves the best performance in the remaining indices (i.e., element quality, min and max angles, and stretch), whereas Blossom-Quad has the lowest quality and is only slightly better than Pave at having fewer triangles that are not expected. It is common for indirect methods (i.e., Blossom-Quad) to have suboptimal performance because of the dependence on prior triangulation. The computational complexities for the three methods are all $O(n^2)$ \citep{pan_huang_wang_cheng_zeng_2021}.

\begin{table}[h!]
    \centering
     \begin{adjustbox}{width=\textwidth}
\begin{threeparttable}
    \caption{Quantitative measurement of the meshing performance of the three methods. All the quality metrics are averaged over the three domains (i.e., domains 7-9).}
    \label{tab:averagaed_mesh_quality}
    \begin{tabular}{cccc}
   \toprule[1pt]
   Metrics & Blossom-Quad & Pave & FreeMesh-RL   \\ \hline
   Singularity (L)  &  388 $\pm$ 209.50 & 146.70 $\pm$ 51.50 & \textbf{132 $\pm$ 50} \\
   Element quality (H)  &  0.72 $\pm$ 0.12 & \textbf{0.79 $\pm$ 0.12} & 0.79 $\pm$ 0.13  \\
   $\vert MinAngle - 90^{\circ}\vert$ (L) & 6.55 $\pm$ 6.91 & \textbf{3.69 $\pm$ 4.60} & 4.02 $\pm$ 5.10 \\
   $\vert MaxAngle - 90^{\circ}\vert$ (L) & 22.16 $\pm$ 11.14 & \textbf{15.69 $\pm$ 14.71} & 15.73 $\pm$ 12.48 \\
   Scaled jacobian (H) & 0.91 $\pm$ 0.08 & 0.94 $\pm$ 0.13 & \textbf{0.94 $\pm$ 0.10} \\
   Stretch (H) & 0.79 $\pm$ 0.08 & \textbf{0.84 $\pm$ 0.10} & 0.83 $\pm$ 0.11 \\
   Taper (L) & 0.15 $\pm$ 0.11 & 0.12 $\pm$ 0.14 & \textbf{0.11 $\pm$ 0.11} \\
   \#Triangle (L) & 2.70 $\pm$ 2.50 & 8 $\pm$ 2.80 & \textbf{0 $\pm$ 0} \\\bottomrule[1pt] 
\end{tabular}
\begin{tablenotes}
      \small
      \item L and H indicate whether the lower value or higher value is preferred, respectively. The value in bold means the best among other approaches in that specific metric.
    \end{tablenotes}
    \end{threeparttable}
\end{adjustbox}
\end{table}

\begin{figure}
    \centering
    \includegraphics[width=\textwidth]{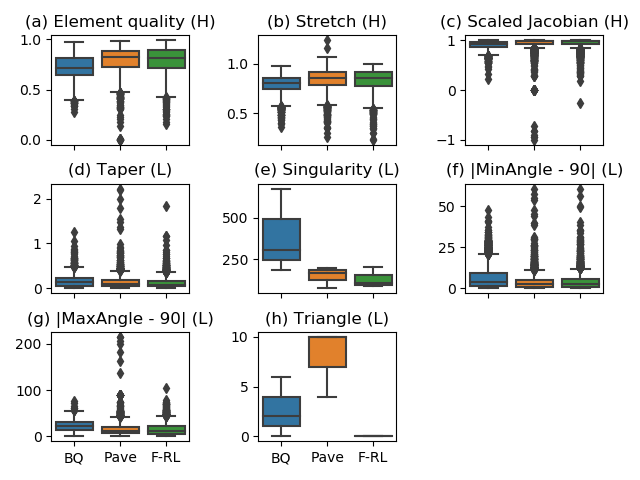}
    \caption{Meshing performance comparison results over eight types of quality indices. BQ represents the Blossom-Quad method; F-RL represents the FreeMesh-RL method. L, H indicate if the lower value or higher value is preferred, respectively.}
    \label{fig:methods comparison}
\end{figure}

The boxplots of the quality metrics are compared in Fig. \ref{fig:methods comparison}, which shows the steadiness and extremely poor situations in each quality measure. Pave has the most outliers in all the quality metrics except singularity and number of triangles, indicating that more elements exist with extremely low quality than others, although it maintains the optimal averaged performance in some metrics. Although Blossom-Quad has the lowest average performance, its generated meshes have the steadiest quality. FreeMesh-RL is in the middle regarding quality steadiness, generating less extreme low quality elements than Pave and maintaining high averaged performance. The more unstable and poor situations require more postprocessing operations.

We also analyzed the knowledge dependence during the algorithm development, as shown in Table \ref{tab:Algorithm development}. The development of the meshing method can be briefly divided into three stages: preprocessing, element generation, and postprocessing. Since Blossom-Quad is an indirect quadrilateral mesh generation method, it requires the preprocessing of triangulation, which is unnecessary for the other two methods. To generate elements, the development of Blossom-Quad requires more complex geometric knowledge than Pave and FreeMesh-RL (see details in \citep{pan_huang_wang_cheng_zeng_2021}). Pave, however, relies on considerable heuristic knowledge, which is time-consuming and inefficient for designers. FreeMesh-RL is the simplest method and only demands basic geometry concepts. Extra treatments such as clean-ups are necessary for Blossom-Quad and Pave, whereas FreeMesh-RL is a clean-up free method.

\begin{table}[h!]
    \centering
     \begin{adjustbox}{width=\textwidth}
\begin{threeparttable}
    \caption{Comparison of knowledge dependence of the three methods during algorithm development. The algorithm development procedure is divided into three stages: preprocessing, element generation, and postprocessing.}
    \label{tab:Algorithm development}
    \begin{tabular}{cccc}
   \toprule[1pt]
   Steps & Blossom-Quad & Pave & FreeMesh-RL   \\ \hline
   Preprocessing  &  Delaunay triangulation & \NA  &  \NA\\
   Element generation  &  Complex geometry knowledge & Heuristic knowledge & Basic geometry concepts\\
   Postprocessing & Clean-ups & Clean-ups & \NA \\\bottomrule[1pt] 
\end{tabular}
\begin{tablenotes}
      \small
      \item 
    \end{tablenotes}
    \end{threeparttable}
\end{adjustbox}
\end{table}

\section{Discussion}
\label{sec: discussion}

Herein, we present a fully automatic quadrilateral mesh generation algorithm, FreeMesh-RL, by using SAC reinforcement learning. With minimal knowledge inputs, this algorithm automatically learns mesh generation policy and performs with comparable quality to commercial systems that conduct quadrilateral mesh generation. This section will discuss the factors behind the performance of the proposed algorithm.

\subsection*{What are the big challenges in mesh generation?}
The big challenges in mesh generation come from two aspects. (1) In real engineering problems, 
the geometric domains are very complex, having various shapes, diverse scales, and different mesh requirements. (2) In mesh generation tools, the rules and knowledge manually acquired are insufficient to tackle all of the situations in a mesh generation problem, especially for large and complex geometries. For example, postprocessing is mandatory to guarantee good quality mesh in Pave and Blossom-Quad, the two representative state-of-the-art systems that we discussed in the last section. The fundamental problem is how to make a mesh generation algorithm general enough to handle various geometric domains with different engineering requirements.

\subsection*{How does FreeMesh-RL tackle the challenges?}
To tackle the challenges in mesh generation, particularly to achieve generalizability for arbitrarily shaped geometries, \citet{zeng_knowledge-based_1993} first proposed an element-wise approach by using a recursive algorithm to generate one element at a time with three primitive rules shown in Figure 5. This element-wise approach exhibits a similar problem structure in reinforcement learning. As such, we formulated the mesh generation problem as an RL problem \citep{pan_huang_wang_cheng_zeng_2021}, as illustrated in Figure 3. Generally, RL is a machine learning paradigm whose success essentially depends on two aspects: (1) the domain-dependent problem formulation, that is, how to represent the environment, states, actions, and rewards; (2) a suitable RL method. This article presents our work on (1) how to optimally design the representation of states, actions, and rewards; (2) how to apply SAC RL to solve the mesh generation problem effectively.

The problem formulation needs insightful abstraction. The three primitive element extraction rules proposed in \citet{zeng_knowledge-based_1993} shown in Figure 5 lays a foundation for the problem formulation. The use of reference vertex makes it possible to have a general representation of the problem, avoiding to deal different shapes and dynamically changing boundaries. We use a partial observation of the environment as the state, thus focusing on the most relevant information of the environment for actions. The reward function design plays a critical role in mesh quality and optimal trade-off of the current reward (the quality of the element to generate) and long-term return (the overall mesh quality). A general observation is that it is critical to make the information represented by states well cover the factors of rewards. 

With respect to RL methods, we select SAC mainly for two reasons. First, its off-policy mechanism allows us to reuse previous experience, thus reducing sampling complexity and making learning more efficient. Secondly and importantly, its distinguishing feature of maximizing policy randomness while maintaining policy performance offers a natural mechanism for an effective trade-off between exploration and exploitation in policy search. This feature of the stochastic policy is particularly beneficial for learning in partially observable environments. 

Our problem formulation, together with the SAC RL algorithm, has demonstrated strong generalizability.
As presented in last section, the model trained with a single domain (see Fig. \ref{fig:training domain comparison} (a)) can mesh various other unseen domains, as shown in Table \ref{tbl:density_comaprison}, Table \ref{tbl:meshing_results_comaprison}, and Fig. \ref{fig:domain generic}. 

The major advantages of the proposed FreeMesh-RL algorithm lie in two aspects: (1) it can automatically acquire and refine knowledge for mesh generation, which is generally labor-intensive and time-consuming, and (2) the automatic knowledge acquisition process does not need predefined samples and labeled data. These advantages effectively tackle some practical challenges in commercial mesh generation software systems.

\subsection*{What is the distinguishing feature of this work?}
The presented research demonstrates how we achieve the integrated intelligence exhibited by the strong generalizability in mesh generation by combining SAC reinforcement learning with the rule-based knowledge representation for states, actions, and rewards. 

In \citet{zeng_knowledge-based_1993}, the authors developed a rule-based knowledge system for mesh generation. 
They created three primitive element extraction rules, which provide a framework for mesh generation problem representation. However, to deal with various boundaries, many rules are needed, but knowledge acquisition is a bottleneck. 
To address this issue, \citet{yao_ann-based_2005} attempted to acquire knowledge for element extraction with MLP neural networks. They introduced the use of reference vertex, which makes the mesh generation problem able to be represented in the same relative space, thus being a MIMO mapping and suitable for using MLP neural networks. Since MLP is a supervised machine learning model, the availability of a large number of samples became a new bottleneck.
To solve this problem, we formulated mesh generation as an RL problem in \citep{pan_huang_wang_cheng_zeng_2021}, took RL as a mechanism to generate meshing samples, selected good quality samples with rules, then fed the selected samples to MLP for training.

In the present article, we further integrate the rule-based knowledge for representing states, actions, and rewards into the framework of SAC reinforcement learning and realize a fully automatic mesh generation system. In this system, rules govern the location of reference vertex, the construction of states, and the quality evaluation of the generated element and the rest of the shape. SAC does the job of exploring possible actions to generate a new element in the current shape, learning to evaluate the quality of the actions, and learning the optimal policy to generate a new element based on the current shape of the geometric domain to mesh. 

\subsection*{What's next?}
As observed in the last section, the performance of the SAC RL algorithm varies with not only its hyperparameters but also the problem formulation and representation and the associated parameter settings. On the other hand, 2D mesh generation is a visible and easily understood problem. We believe this study of automatic 2D mesh generation with RL can be developed into an RL testbed for RL research.

FreeMesh-RL is currently limited to 2D domains.
We will apply the proposed framework to 3D mesh generation by reformulating the state representation and a few primitive actions for hexahedron elements as well as the reward function. 

In the fourth industrial revolution, engineering is transforming into digital engineering \citep{JIDPS-V25N1-AI4DE, Huang2022}, where digital data and models will be shared across the engineering lifecycle. This will make it possible to collect a large volume of meshing examples. Currently, our model is trained with just a single simple geometric domain and shows the capability to mesh in comparable quality to representative commercial software. In real-world engineering for 3D meshing, there will be various challenging geometric domains and meshing requirements for different purposes. We intend to leverage transfer learning \citep{1976-transfer-learning, 2021-transfer-learning-survey} and the available big data of real-world mesh samples to develop deep neural networks, which will be pretrained with excellent samples and can be repurposed by further training with a few samples for a specific task with specific meshing requirements.

\section{Conclusion}
\label{sec: conclusion}
This article presented our research on applying soft actor-critic reinforcement learning for automatic mesh generation. We designed and implemented FreeMesh-RL, a fully automatic quadrilateral mesh generation algorithm with SAC reinforcement learning. With minimal knowledge inputs, this algorithm automatically learns mesh generation policy and performs with comparable quality to commercial systems that conduct quadrilateral mesh generation. Further research can go in several directions. We will extend our current 2D automatic mesh generation algorithm with RL for 3D mesh generation. Also, based on this study of automatic 2D mesh generation with SAC RL, we will develop it into an RL testbed for RL research.

\section*{Acknowledgment}

The support of the NSERC Discovery Grant (RGPIN-2019-07048) is gratefully acknowledged.

\bibliographystyle{elsarticle-harv} 
\bibliography{main}





\end{document}